\documentclass[journal]{IEEEtai}

\usepackage[colorlinks,urlcolor=blue,linkcolor=blue,citecolor=blue]{hyperref}

\usepackage{color,array}

\usepackage{graphicx}
\usepackage[]{algorithm2e}
\usepackage{subcaption}
\usepackage{multirow}
 \usepackage{amssymb}
\usepackage{float}
\RestyleAlgo{ruled}
 \usepackage{amsmath}
\usepackage{comment}
\usepackage{caption}
\usepackage{cite}
\usepackage{colortbl}

\begin{document}

\title{FaceTopoNet: Facial Expression Recognition using Face Topology Learning}

\author{Mojtaba Kolahdouzi, Alireza Sepas-Moghaddam, \IEEEmembership{Member, IEEE}, and Ali Etemad, \IEEEmembership{Senior Member, IEEE}
\thanks{This work was funded by Irdeto Canada Corporation and the Natural
Sciences and Engineering Research Council of Canada (NSERC).

Mojtaba Kolahdouzi is a Ph.D. student in the Department of Electrical and Computer Engineering and Ingenuity Labs Research Institute at Queen’s University, Kingston, Canada
(19mk73@queensu.ca)

Alireza Sepas-Moghaddam is a research collaborator in the Department of Electrical and Computer Engineering and Ingenuity Labs Research Institute at Queen’s University, Kingston, Canada
(alireza.sepasmoghaddam@queensu.ca)

Ali Etemad is an Associate Professor and a Mitchell Professor in AI for Human Sensing and Understanding, in the Department of Electrical and Computer Engineering and Ingenuity Labs Research Institute, at Queen’s University, Canada. (e-mail: ali.etemad@queensu.ca) }
% \thanks{The next few paragraphs should contain the authors' current affiliations, including current address and e-mail. For example, F. A. Author is with the National Institute of Standards and Technology, Boulder, CO 80305 USA (e-mail: author@boulder.nist.gov).}
% \thanks{S. B. Author, Jr., was with Rice University, Houston, TX 77005 USA. He is now with the Department of Physics, Colorado State University, Fort Collins, CO 80523 USA (e-mail: author@lamar.colostate.edu).}
% \thanks{T. C. Author is with the Electrical Engineering Department, University of Colorado, Boulder, CO 80309 USA, on leave from the National Research Institute for Metals, Tsukuba, Japan (e-mail: author@nrim.go.jp).}
% \thanks{This paragraph will include the Associate Editor who handled your paper.}
}

% \markboth{IEEE Transactions on Artificial Intelligence}
% {FaceTopoNet: Facial Expression Recognition using Face Topology Learning}

\maketitle

\begin{abstract}
Prior work has shown that the order in which different components of the face are learned using a sequential learner can play an important role in the performance of facial expression recognition systems. We propose FaceTopoNet, an end-to-end deep model for facial expression recognition, which is capable of learning an effective tree topology of the face. Our model then traverses the learned tree to generate a sequence, which is then used to form an embedding to feed a sequential learner. The devised model adopts one stream for learning structure and one stream for learning texture. The structure stream focuses on the positions of the facial landmarks, while the main focus of the texture stream is on the patches around the landmarks to learn textural information. We then fuse the outputs of the two streams by utilizing an effective attention-based fusion strategy. We perform extensive experiments on four large-scale in-the-wild facial expression datasets - namely AffectNet, FER2013, ExpW, and RAF-DB - and one lab-controlled dataset (CK+) to evaluate our approach. FaceTopoNet achieves state-of-the-art performance on three of the five datasets and obtains competitive results on the other two datasets. We also perform rigorous ablation and sensitivity experiments to evaluate the impact of different components and parameters in our model. Lastly, we perform robustness experiments and demonstrate that FaceTopoNet is more robust against occlusions in comparison to other leading methods in the area.
% The results indicate that the proposed method outperforms other facial expression recognition methods.
\end{abstract}

\begin{IEEEImpStatement}
Systems that perform facial expression recognition (FER) have a wide variety of applications in real-world problems such as human-machine interaction, robotics, mental health management systems, to name but a few. Our proposed solution achieves better recognition rates in comparison to many other methods in the field of FER, allowing for development of more accurate systems. Moreover, our proposed method demonstrates more robustness against occlusions, which can be beneficial for in-the-wild scenarios. Lastly, our model can learn trees that are optimized, yet also generalize well across different datasets and perform considerably better than human-designed face trees.

% understanding human behavior, emotion recognition, pain intensity detection, and robotics. 
% Because our devised method sets new state-of-the-art recognition rate, it can provide more accurate expression recognition in any of the applications, which in turn can increase their efficiency. 

% which can improve the outcome of facial expression recognition in scenarios in which occlusions usually exist, like 
% the relationships of different components of a face. This can provide more transparency as well as insight into the applications which are decision-critical, like pain intensity detection in humans.

% Nowadays, because important components of the face like mouth contribute to different expressions, deep learning models intending to do expression recognition are more focused on part-based methods, i.e., graph-based methods. Despite being successful, graph-based methods for FER do not have a consensus on the topology of the underlying graph. Our proposed method, named FaceTopoNet, tries to overcome this issue by learning a tree, rather than assuming a pre-defined or engineered tree. FaceTopoNet not only achieves the highest recognition rate in AffectNet, FER2013, and ExpW datasets compared to current state-of-the-art, but also it shows more robustness against different occlusions.

\end{IEEEImpStatement}


\section{Introduction}

\IEEEPARstart{F}{acial} expressions play a key role in conveying a wide range of human emotions~\cite{jaques2016understanding, picard2000affective}. Thus, considerable attention has been given to automatic facial expression recognition (FER) using machine learning and deep learning methods, where the goal is to identify different expressions from images or videos of the face~\cite{sepas2021capsfield, shuvendu2021, sepas2020facial}. Recent progress in deep neural networks and computational resources, along with the availability of vast amounts of data, have resulted in great progress in the area of FER~\cite{sepas2019deep}. Applications of FER include but are not limited to emotion classification~\cite{shuvendu2021}, pain intensity detection~\cite{bargshady2020enhanced}, age estimation~\cite{pei2019attended}, estimation of the user's experience during consumption of multimedia content~\cite{porcu2020estimation}, and detection and prediction of the severity of depression~\cite{li2020depression}, among many others. A detailed recent review of FER is presented in~\cite{surveyAsli}.

\begin{figure}[!t]
    \begin{center}
     \includegraphics[scale=0.16]{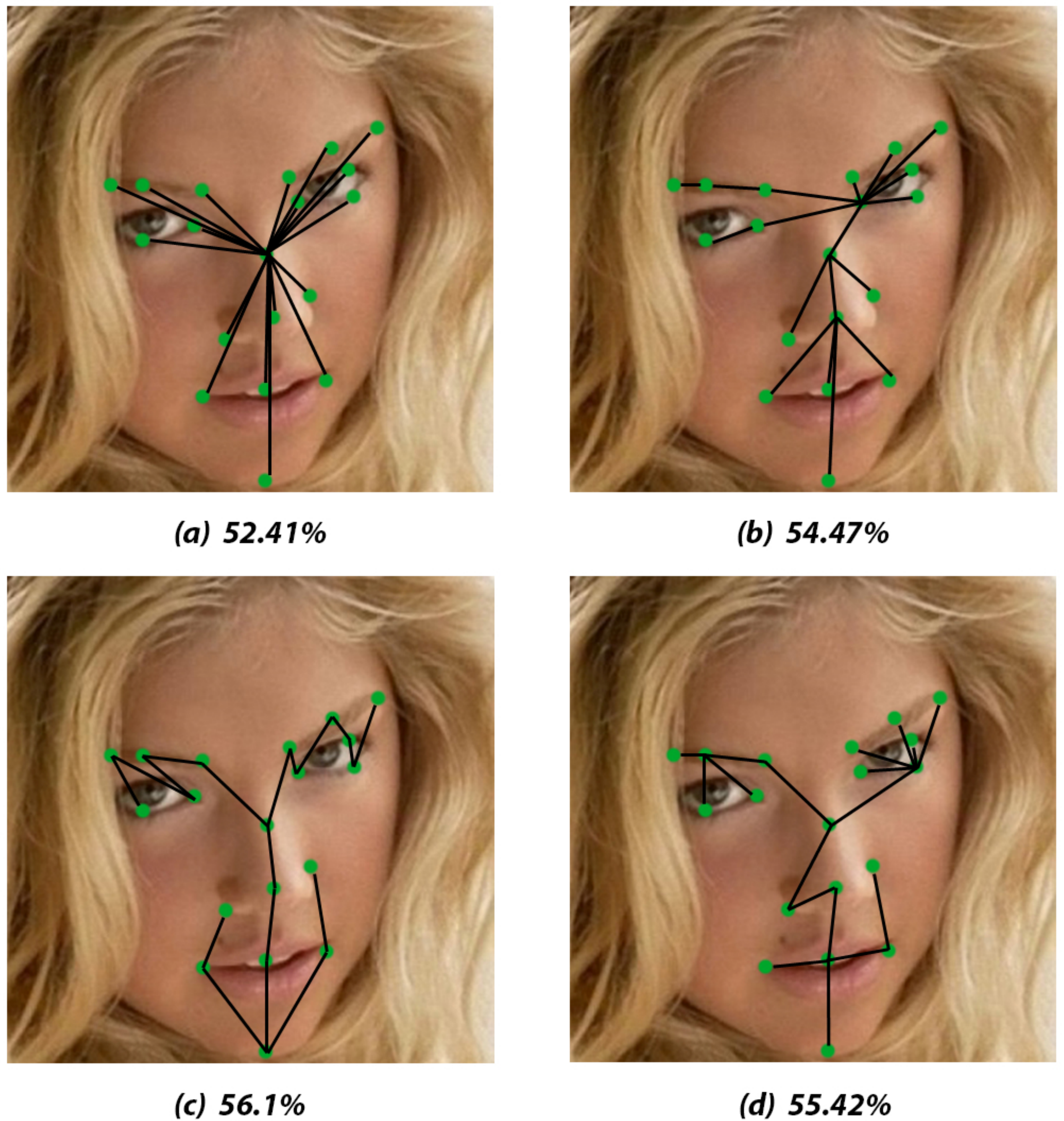}
    \end{center}
\caption{A face image together with 4 random trees. Associated recognition rates are shown beneath each image.}
\label{fig:banner}
\end{figure}

While many early deep FER approaches such as~\cite{jung2015joint} and ~\cite{ghimire2017facial} focused on \textit{holistic} representations of the face, many newer methods have shifted focus to \textit{part-based} approaches to describe the face as distinct~\cite{happy2020expression} or disentangled components~\cite{liu2019hard}. The reason for this focus toward part-based methods is that certain regions of the face such as mouth or periocular areas have been shown to play different roles in reflecting different expressions~\cite{yu2020facial}. Part-based FER techniques typically analyze face regions separately to extract local features, which are then merged to produce global representations~\cite{happy2020expression}. As a result, given that graph-based methods generally lend themselves well to part-based problems, they have recently gained popularity for FER~\cite{GCN1,GCN2,GRNN, G1, G2, G3}. In such approaches, each node in a graph can be associated with a facial region, where an edge can reflect the connections or relationships between the regions, for instance their joint movement or distance.

Despite the fact that a number of graph-based FER methods have shown promising results~\cite{liu2019facial, GCN2}, the underlying graph topology used to learn the structure of the face is not generally agreed upon. In other words, there is no single way of knowing how various face regions should be connected such that the model's performance would be maximized. Graph-based FER methods assume the topology of underlying graphs as being either fully-connected~\cite{GRNN, GCN2,G2} or partially-connected~\cite{G1,G3,GCN1}. In the fully-connected scenario, one must add ${n \choose 2}$ edges to the underlying graph, which makes the training process cumbersome. 
Additionally, some of the edges are unnecessary within the graph, which may cause overfitting. In the partially-connected scenario, in light of the fact that the topology is pre-defined and engineered as opposed to being learned, it may not include the optimal or near-to-optimal edges between face regions. In any case, the underlying graph topology can significantly impact the performance of any subsequent learner that adopts the information obtained from different facial regions toward part-based FER. As an example Figure~\ref{fig:banner} depicts a face image together with 4 random trees which have been used to connect different regions of the face. Traversal of the trees yields 4 distinct sequence-based embeddings, using which we can conduct a 7-class FER task using 4 independent sequential learners (long short-term memory (LSTM) has been used in this case). The associated recognition rates (RR) are provided beneath each image in the Figure~\ref{fig:banner}, where we observe a 3.7\% variance, demonstrating the necessity of selecting the optimal tree topology.

In this paper, we present an end-to-end pipeline, FaceTopoNet, that learns an effective tree of the face for robust FER in a deep neural architecture.
First, we assume a weighted fully-connected graph obtained from facial landmarks, which is then optimized by an evolutionary algorithm~\cite{li2011cooperatively}. This graph is then used to form a minimum-cost spanning tree, the traversal of which forms a sequence of facial nodes. This sequence is then used for two subsequent streams, which we name \textit{texture} and \textit{structure} streams. In the structure stream, the obtained sequence is used to feed a sequential learner followed by soft attention to focus on the salient cells of the learner. In parallel, in the texture stream, we apply local patches around each of the extracted nodes and encode each region using a simple ConvNet. These encoded representations are used to feed a sequential learner similar to the structure stream, followed by soft attention. The final embeddings of the two streams are then fused for final expression recognition. To the best of our knowledge, our work is the first to optimize the relevant order of facial regions for sequential learning in the context of FER. We perform rigorous experiments on four in-the-wild large-scale FER datasets, AffectNet~\cite{Affectnet}, FER2013~\cite{FER2013}, RAF-DB~\cite{li2017reliable}, and ExpW~\cite{SOCIALRELATION_2017}, and one lab-controlled dataset, i.e. CK+, to illustrate the efficacy of our devised method. Our method outperforms the state-of-the-art results on three of the datasets and achieves competitive results on the other two. We further perform additional experiments on ablations and even evaluate the robustness of our method to occlusions in different key regions of the face.

our contributions are three-fold. (\textbf{1}) We propose a novel end-to-end deep architecture to identify and use the effective tree topologies for FER. 
% of face trees for finding the effective order of the embeddings to be fed to two sequential learners located in structure and texture streams of our pipeline.
These face trees are found by using the idea of weighted complete graphs and minimum-cost spanning trees. They are then used to determine the input order of the embeddings to be fed to two sequential learners located in two separate streams responsible for learning the structure and texture of the face.
(\textbf{2}) Our experiments demonstrate the strong performance of our method by setting new state-of-the-art results on AffectNet, FER2013, and ExpW, and performs competitively on RAF-DB and CK+. (\textbf{3}) Our solution is robust to different occlusions added to input images. Further, we find that our method automatically adapts and finds different face trees for each dataset, and uses them to obtain improved performance.
It is worth mentioning that in traditional landmark-based methods the input order of the sequential learner is not learned as opposed to FaceTopoNet. In fact, FaceTopoNet provides a way to simultaneously learn the input orderings of a sequential learner and its free parameters.

This paper is an extension of our prior work~\cite{kolahdouzi2021face} which has been published in FG 2021 as a short paper. In comparison, this paper adds the followings: (\textit{i}) We use three additional publicly available datasets, namely RAF-DB, ExpW, and CK+, to evaluate the proposed method. (\textit{ii}) We analyze and discuss the effects of using different sequential learners and encoders with different pre-trainings. (\textit{iii}) We perform robustness experiments toward occlusions against other state-of-the-art methods by manually removing different key regions of the face. (\textit{iv}) We study the effect of different patch sizes which are located around the facial landmarks on the overall recognition rates. (\textit{v}) The effects of the number of extracted facial landmarks is studied (\textit{vi}) We perform a cross-dataset study utilizing face trees. (\textit{vii}) Lastly, we perform a direct comparison between the \textit{learned} face trees and prior human-defined trees used for FER.

The rest of this paper is organized as follows: Section~\ref{sec:related-work} provides a summary of the related work in the field. The proposed method including the detailed description of tree topology learning is given in section~\ref{sec:proposed-method}. Section~\ref{sec:experiments} presents the results of the extensive experiments along with the detailed ablation analysis, and finally section~\ref{sec:conclusion} concludes the paper.

%%%%%%%%%%%%%%%%%%%%%%%%%%%%%%%%%%%%%%%%%%%%%%%%%%%%%%%%%%%%%%%%%%%%%%%%%%%%%%%%%%%%%%%

\section{Related Work} \label{sec:related-work}
FER systems can be divided into static and dynamic categories with respect to the type of input data~\cite{surveyAsli}. In static-based methods, the facial features are encoded based on the spatial information available in a \textit{single} image, whereas dynamic-based methods additionally exploit the temporal relationships between adjacent frames available in the input \textit{video}. Our focus in this paper lies in the first category. Hence, we review relevant papers for image-based FER in this section. Additionally, given the use of graphs in our method, we provide a review of the available techniques that incorporate graph theory in FER systems.

\subsection{FER with Deep Learning}

In one of the early neural network-based methods for FER, a shallow CNN was used for feature learning, followed by a linear support vector machine (SVM) for classification~\cite{tang2013deep}. Another method, HoloNet, was introduced in~\cite{yao2016holonet}, combining residual structures with concatenated rectified linear activation scheme ~\cite{shang2016understanding}. In addition, an inception-residual structure was adopted in the final layers to enable multi-scale feature extraction. In~\cite{sang2017facial}, a nine-layer CNN with a multi-class SVM classifier was used for FER.

In~\cite{liu2019hard}, the notion of disentanglement was used to make FER systems more robust against pose, illumination, and other variations. In this context, a radial metric learning strategy was proposed for disentangling identity-specific factors from facial input images. In~\cite{chen2019facial}, an attention method was proposed to selectively focus on facial muscle moving regions. To guide the attention learning process, prior domain knowledge was incorporated by considering the differences between neutral and expressive faces during training. In~\cite{wang2020region}, another attention-based method was suggested. The method used a region-attention network to identify the importance of different facial regions for FER, thus assigning higher weights to facial regions with greater importance. This resulted in further robustness to occlusions and pose variations.

In~\cite{jiang2020accurate}, the so called advanced softmax loss was proposed to reduce the negative effects caused by data imbalance, particularly when the classifier is biased toward the majority class. In~\cite{kollias2020deep}, the authors synthesized facial affects from a neutral image for data augmentation in FER systems. Their model accepts a neutral image and a basic facial expression as inputs and fits a 3D Morphable model~\cite{booth20173d} to the neutral image, deforming it for adding the expression. A model called MA-Net was proposed in~\cite{zhao2021learning} for conducting in-the-wild FER. The proposed solution includes three components. In the first component, named feature pre-extractor, mid-level features of face images are extracted. The second component tries to fuse different features which are related to different receptive fields, and finally in the third component, the network is guided to be more focused on the more important local features. A model named TransFER was proposed in~\cite{xue2021transfer} for FER. The approach is capable of learning complex local representations and is comprised of three sub-networks: (\textit{i}) multi-attention dropping, forcing the model to come up with more diverse local patches; (\textit{ii}) vision transformer to model the relations existing among different local patches; and (\textit{iii}) multi-head self-attention dropping which drops a self-attention module randomly to make the model learn different local patches in a diverse way.

Since facial attributes which are associated with identity, like race or age, can hinder the efficacy of FER systems, an identity-free FER system is proposed in~\cite{cai2021identity}. The authors have made use of a generative adversarial network to transform a given face image with a specific identity to its ``average'' identity version while preserving the same expression. A new way to view the facial expressions was proposed in~\cite{ruan2021feature}.  
% Specifically, the authors have assumed that the information of an expression can be modeled as a shared plus a unique information, and thus 
Specifically, the authors utilized a feature decomposition network to extract shared information among expressions and a feature reconstruction network to extract expression-specific information. In~\cite{huang2021identity}, for data augmentation purpose, the authors used StarGAN~\cite{choi2018stargan} to synthesize identity-specific face images with basic emotions. Then, a deep network was used to extract latent embeddings from both synthesized and real images. These embeddings were used to train a Mahalanobis metric network for performing FER. In~\cite{bisogni2022impact}, the authors design a FER system to assist healthcare systems. In this regard, they propose different convolutional neural network architectures to handle multiresolution facial images. Hossain \textit{et al.} in~\cite{hossain2021unified} propose a three-component FER system. First component includes the detection of facial regions, the second component uses CNNs to extract the embeddings, and the third component adopts different techniques like transfer learning to extract more distinctive embeddings. A novel data augmentation technique is introduced in~\cite{umer2022facial} for performing FER. This technique includes applying edge enhancement filters like bilateral filtering on the images. Although deep learning methods provide the state-of-the-art results in FER, some papers like~\cite{oztel2018ifer} use a conventional machine learning approach (feature extraction, feature classification, and classification) to perform FER and achieved reasonable results.

In~\cite{hasani2019bounded}, a novel residual-based CNN architecture, Bounded Residual Gradient Network (BReG-Net), was proposed for FER. This CNN architecture includes 39 layers of residual blocks, where the standard skip-connections in the ResNet blocks were replaced by a differentiable function with a bounded gradient, thus avoiding the gradient vanishing and exploding problems. As an extension to~\cite{hasani2019bounded}, the same authors proposed BReG-NeXt in~\cite{hasani2020breg}, which includes more trainable parameters and offers more flexibility. A trainable parameter has also been added to the residual units. While this complex mapping adds few more parameters to each residual block, it extracts more representative features in comparison to its counterparts like ResNet-50. 
% improving fitting and training of the network and thus better recognition of facial expressions.

\subsection{Applications of Graphs in FER}

Neural networks on graph topologies have recently gained momentum for visual analysis tasks~\cite{VQA,ICLR}, including FER~\cite{GCN1,GCN2,GRNN, G1}. In~\cite{GRNN}, Gabor features were first extracted from texture information around 49 facial landmarks and were then combined with the Euclidean distances calculated between landmark pairs to feed a bidirectional gated recurrent unit. In~\cite{GCN2}, local regions around five landmarks, including center of left and right eyes, nose, and left and right mouth corners, were extracted to be then learned using a GCN. Similarly in~\cite{G2} 28 landmarks corresponding to eyebrows and mouth were selected to create three separate graphs for left/right eyebrows and mouth. These graphs were then processed using a graph temporal convolutional network. In~\cite{G1}, 34 landmarks, corresponding to three facial components including left eye and eyebrow, right eye and eyebrow, and mouth, were first detected and the connections between these components were established with respect to their psychological semantic relationships. Histogram of orientation features and the XY coordinates of the nodes were then combined to feed a spatial-temporal semantic graph network. A novel multi-task learning was proposed in~\cite{antoniadis2021exploiting}, which adopted GCNs to model the dependencies among the discrete and continuous models of emotions. To be more specific, a shared representation was learned for both discrete and continuous models of emotion, and the final classifier of the discrete model and the regressor of the continuous model were jointly learned through a GCN. In~\cite{liu2021video}, the authors adopted a GCN into a well-known CNN-RNN based model for performing FER. The role of the GCN was to help the model learn more salient features. Additionally, an LSTM layer was utilized to learn long dependencies between the GCN-extracted features.  

Unlike all these techniques which employ fully connected graphs for FER, a few solutions such as~\cite{G1} and~\cite{GCN1} assume a pre-defined subset of connections between the nodes followed by a variation of graph convolutional networks. Nonetheless, to the best of our knowledge, all the graph-based FER solutions utilize a pre-defined graph topology (whether fully connected or otherwise) to be then used for every scenario. The problem with these pre-defined graphs is that they do not necessarily include the optimal set of connections, and once designed, stay unchanged for different datasets.  We aim to tackle this problem by proposing a novel pipeline in which a unique effective tree topology can be learned given the problem space.

%%%%%%%%%%%%%%%%%%%%%%%%%%%%%%%%%%%%%%%%%%%%%%%%%%%%%%%%%%%%%%%%%%%%%%%%%%%%%%%%%%%%%%%%%%%%%%%

\section{Proposed Method} \label{sec:proposed-method}
\subsection{Problem Setup and Solution Overview}
%The topology of facial components and their relationships can be modeled in the form of a graph, thus providing better representations compared to holistic approaches for facial image analysis tasks~\cite{Zhou2020facial}. Such graphs consist of vertices corresponding to facial landmarks connected through edges, where each edge represents the relationship between two landmarks~\cite{jiang2017emotion}.
Graphs can be used to model the topology and the relationship between different facial landmarks. Each vertex in these graphs can represent a facial landmark, while the existence of an edge between two vertices (facial landmarks) would consequently demonstrate that there is a relationship between the two. In~\cite{Zhou2020facial}, it was shown that these graphs provide a better representation when compared to holistic approaches.
The constructed graphs can then be processed using deep networks like 3D-CNNs~\cite{liu2014deeply}, GCNs~\cite{liu2020video}, or RNNs~\cite{GRNN} (which we refer to as \textit{sequential learners})~\cite{cho2014learning}. 
%the selection of graph topology can have a considerable effect on the overal performance of the subsequent sequential learner.
As shown in Figure~\ref{fig:banner}, when using a sequential learner, the accuracy of a solution can be affected by the topology of the graph being processed.
%As evident from Figure~\ref{fig:banner},the design of the graph topology can have a considerable effect on the overall recognition rate of a solution. 
Nonetheless, to the best of our knowledge, no technique for learning the topology of %face 
graphs for FER has been previously proposed in the literature.

%To fill the aforementioned gap, we propose a novel method in this paper to learn the effective face tree, thus effectively modeling 
%these relationships for emotion recognition. 
To address the aforementioned problem, we propose a new solution in this paper to learn the topology of the graph in the form a face tree, and then use this tree to further learn effective representations from the face. 
% from the structure and texture of the face. 
% This learned topology yields higher recognition rates.

\begin{figure*}
    \begin{center}
    \includegraphics[width=\textwidth]{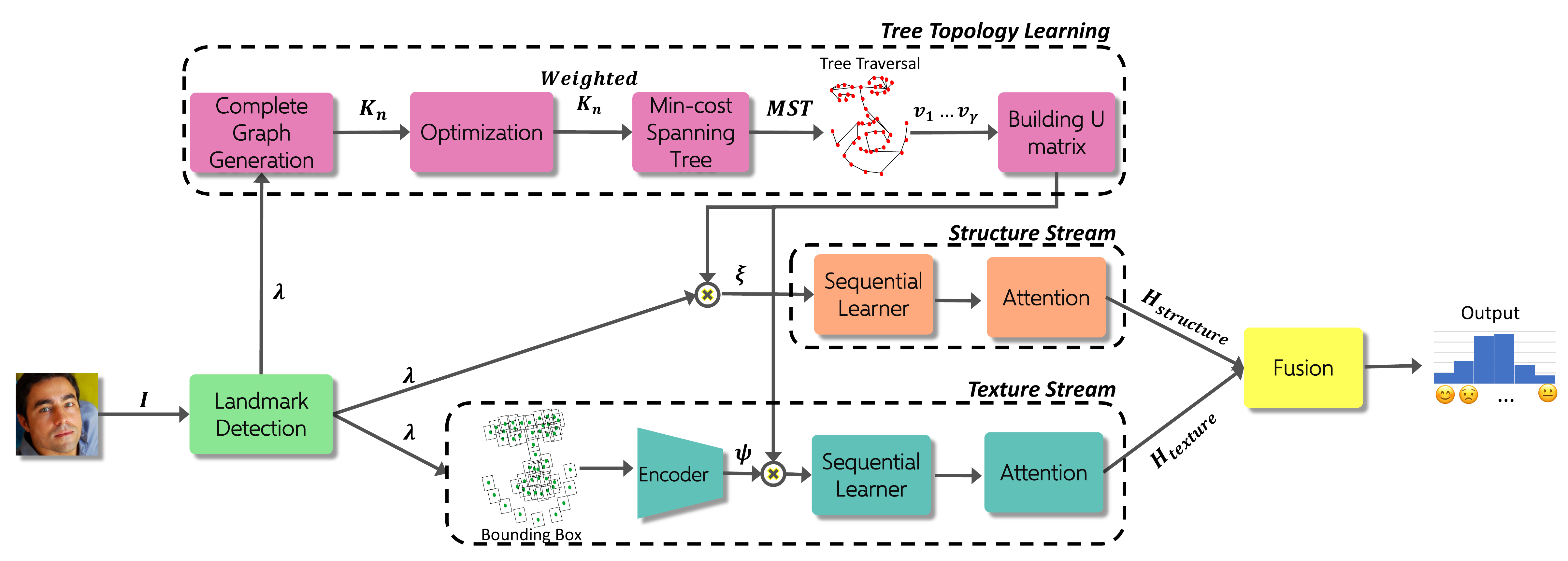}
    \end{center}
\caption{The pipeline of the proposed FaceTopoNet.}
\label{fig:pipeline}
\end{figure*}

%the relationships among different facial components for emotion recognition. After this step, 
In our designed solution, for learning the salient information of the face, two important streams are considered: \textit{structure} and \textit{texture}. The backbone of these two streams are two sequential learners. %Accordingly, our proposed end-to-end method adopts one stream for learning the structure of the face considering the obtained tree topology and another for learning the texture of important patches around facial landmarks, also by considering the same tree. Our proposed method traverses the tree to form a sequence, which is then recurrently learned  using a sequential learner.
Accordingly, our proposed end-to-end method uses the spatial information of the extracted facial landmarks in the structure stream and the texture information of the patches around these facial landmarks in the texture stream. The final embeddings to be fed into the sequential learners are determined by the topology of the learned face tree. In other words, we traverse the face tree to generate a sequence, and this sequence determines the ordering of the embeddings to be fed into the sequential learners.
Our proposed pipeline's architecture is depicted in Figure~\ref{fig:pipeline}. What follows is the detailed description of our tree topology learning procedure along with the two-stream solution.

\subsection{Tree Topology Learning} \label{sec:TreeTopologyLearning}
% we aim to find the optimal topology of a face tree, which will be subsequently utilized by the sequential learners. In this context, 
First, we employ a landmark extractor $\Phi$ to detect facial landmarks from an input image $I$ such that:
\begin{equation}
     {\lambda} = \Phi(I), %{i=1,\cdots,n}
\end{equation}
where  $\lambda = [\lambda_1 \cdots \lambda_i \cdots \lambda_{n}]$, $\lambda_i$ represent the coordinates of the \textit{i}th facial landmark in Euclidean space, and $n$ is the number of extracted facial landmarks. The optimal value of $n$ will be studied in the experiments section. In this work, we use a deep regression architecture~\cite{lv2017deep} as $\Phi()$. Next, we suppose that the extracted facial landmarks are vertices of a graph. We thus have a graph with $n$ vertices where each vertex represents a facial landmark. We then add all the possible edges, obtaining a complete graph $K_{n}$. 
%which includes all the extracted facial landmarks. In this work, we use a deep regression architecture~\cite{lv2017deep} as $\Phi$ and subsequently extract $n$ facial landmarks - the optimal value of n will be addressed in the experiment section. Next, by supposing all $\lambda_i$ as the vertices of a graph, we build a complete graph, $K_{n}$, by adding all the possible edges. 

%Using a mechanism which we describe below in this section, 
Next, an optimization algorithm (described below in this section) assigns weights to the edges of $K_{n}$, making it a weighted complete graph. Following, we extract a minimum-cost spanning tree from $K_{n}$. This minimum-cost spanning tree can be expressed as:
\begin{equation}\label{Eq:MST}
    % MST = \{(V_{MST}, E_{MST})| V_{MST}=V_{K_{n}}, E_{MST} \subset E_K_{n}, \\ \sum_{\forall e_{ij} \in E_{MST}}{e_{ij}} \:is\: minimum, (V_{MST}, E_{MST})\: is\: a\: tree \}.
    MST = \{(V_{MST}, E_{MST})\},
\end{equation}
where $MST$ is the minimum-cost spanning tree, and $V_{MST}$ and $E_{MST}$ represent the vertices and edges of $MST$, respectively. Here, $V_{MST}=V_{K_{n}}$, $E_{MST} \subset E_{K_{n}}$, where $V_{K_{n}}$ and $E_{K_{n}}$ denote the vertices and edges of $K_{n}$. Moreover, in this setup, $\sum_{\forall e_{ij} \in E_{MST}}{e_{ij}}$ is minimum. % and  $(V_{MST}, E_{MST})\: is\: a\: tree$
For finding $MST$, we use Algorithm~\ref{alg:prim} based on ~\cite{jiang2009research}. 
%This algorithm belongs to a family of greedy methods which visits each of the vertices of $k_{n}$ only once. During this visit, it chooses an edge with the minimum value and adds it to the $E_{MST}$ set. 
%It is worth mentioning that the reason behind converting the obtained graph $K_{n}$ to a tree for sequence generation is that trees are acyclic, and are thus better suited for traversal as well as sequence formation.

\begin{figure*}
    \begin{center}
    \includegraphics[width=0.8\textwidth]{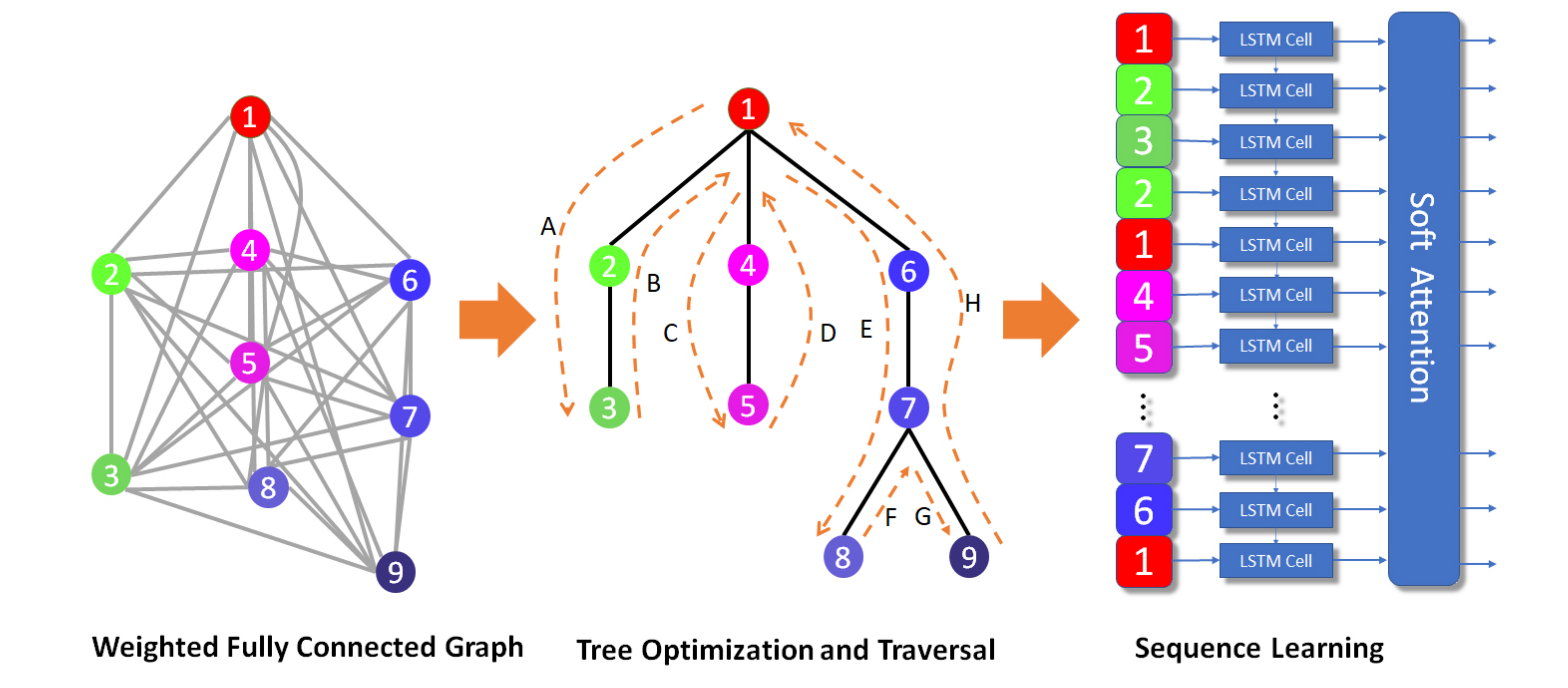}
    \end{center}
\caption{The details of the tree traversal and the following embedding formation. The built minimum-cost spanning tree is traversed by using Preorder depth first order, which produces the sequence 1-2-3-2-1-4-5-4-1-6-7-8-7-9-7-6-1. The resulting embedding of the sequence is fed to a sequential learner.}
\label{fig:TreeTraversal}
\end{figure*}

By utilizing the $MST$, we aim to generate a sequence which determines the ordering of the embeddings for both structure and texture streams. The reason behind converting the complete graph $K_n$ to a tree is that trees are acyclic in nature and are thus better suited for sequence formation. For the purpose of sequence formation, we traverse $MST$ using the preorder depth first algorithm described in~\cite{kozen1992depth}. In the context of our problem, we start the MST traversal from the most centered vertex, i.e., nose tip. 
%Upon reaching a leaf node during the tree traversal, the algorithm backtracks and returns to the previous node. 
Ultimately, a sequence of vertices is built. We depict the tree traversal in Figure~\ref{fig:TreeTraversal}. In this figure, for a clear visualization, we explain the process for a tree with only $n=~$9 nodes instead of the entire set of landmarks. %We show how the method is used to traverse the tree and how it backtracks when reaching a leaf node.

% Suppose that the sequence generated by the traversal operation is denoted by $v_{\alpha}v_{\beta}\cdots v_{\gamma}$ where $v_{\alpha}$ is the vertex (landmark) $\alpha$ of the tree. 

Suppose that the sequence generated after the tree traversal is denoted by $v = [v_1 \cdots v_{\gamma}]$, where $v_1$ is the first vertex (landmark) of the tree, and $\gamma$ is one of the existing vertices between $1$ and $n$. Accordingly, the number of elements in vector $v$ is provably $2n-1$.
%$2(n-1) + 1$.
 By utilizing $v$, we build matrix $U=[u_1 \cdots u_{\gamma}]$, where $u_1$ is a vector whose elements are zeros except for the first element. Similarly, $u_{\gamma}$ is a vector whose elements are zeros except for the $\gamma^{\text{th}}$ element. Matrix $U$ is then used to generate the input embeddings of the sequential learners, which will be described in Section \ref{sec:StructureStream}.

% \textit{Proof:} Every tree with  has $x-1$ vertices. In the pre-order depth first traversal described in figure~\ref{fig:TreeTraversal}, we pass every edge exactly $2$ times. Thus, the generated sequence include $2 \times (V-1) + 1$ vertices
% calculated as follows.
% \textit{Lemma}. The generated sequence described by the traversal operation above includes 
% vertices (landmarks). 

To learn an effective face tree structure, the weights of $K_{n}$ need to be updated. In other words, changing the weights of $K_n$ results in different face trees, thus different sequences. Given the greedy nature of Algorithm~\ref{alg:prim}, it is difficult to express the output of our pipeline as a differentiable function of the weights of $K_{n}$. In dealing with such problems, various optimization algorithms can be used to learn effective trees. In this work, we utilize the Cooperatively Coevolving Particle Swarms for Large Scale Optimization (CCPSO2) algorithm~\cite{li2011cooperatively}, which is an evolutionary method, to optimize the weights of $K_{n}$. CCPSO2 belongs to a family of meta-heuristic cooperative algorithms, which makes use of swarm intelligence as well as their cooperation to estimate the solution of optimization problems of up to 2000 variables effectively\cite{li2011cooperatively}
% . Additionally, CCPSO2 can estimate the solution high-dimensional optimization problems (up to 2000 variables) effectively.
%without getting stuck at local minima due to its reliance on both Normal and longer tailed Cauchy distributions to perform both exploration and exploitation in the problem space~\cite{li2011cooperatively}. 
Due to the fact that our search space in this work has ${n \choose 2}$ dimensions, for a reasonably constrained value such as $n=50$, CCPSO2 can perform effectively for this problem. We use this optimization technique to learn the weights of $K_{n}$ while minimizing the loss function of the model $\mathcal{L}_{total}$, which we will define later in Section \ref{sec:loss}. It is worth mentioning that while CCPSO2 does not promise an optimal solution (it can get stuck in local optima like many other optimizers), what matters is the generalization ability of the converged face tree on unseen data, which will be extensively tested on 5 different FER datasets in the experiment section of the paper.

\begin{algorithm}[t]
\SetAlgoLined
\KwResult{MST}
\textbf{Initialization}: $visited=\{i\}$, $unvisited=\{0,1,..,i-1,i+1,...,{n\choose{2}}\}$, $MST=\varnothing$\\
\While{$unvisited\not=\varnothing$}{
  find edge $e=(P, Q)\, \ni$ $P \in visited $ and $Q \in unvisited $ and $e$ has the smallest weight\\
  $MST = MST \cup \{e\}$\\
  $visited = visited \cup \{Q\}$\\
  $unvisited = unvisited - \{Q\}$\\
 }
 \caption{Minimum cost spanning tree} \label{alg:prim}
\end{algorithm}

% \begin{algorithm}[t]
% \SetAlgoLined
% \KwResult{global\_best}
% \textbf{Initialization}: $r=1225$, $\hat{r}=1$, $R = \{1, 5, 7\}$ \\
% \While{$loop\_counter<=40$}{
%   \If{global\_best does not show any improvement}{
%     choose a random $\hat{r}$ from the set $R$ and set m as $m=1225/\hat{r}$
%   }
%   Build $m$ swarms, each of which has $\hat{r}$ dimensions\\
%   \For{each swarm $\alpha \in \{1 ... m\}$}{
%     \For{each particle $\beta \in \{1 ... swarm\_size\}$}{
%         \If{$f(\textbf{b}(\alpha, W_\alpha.x_\beta)) < f(\textbf{b}(\alpha, W_\alpha.y_\beta))$}{
%             $W_\alpha.y_\beta \gets W_\alpha.x_\beta$
%         }
%         \If{$f(\textbf{b}(\alpha, W_\alpha.y_\beta)) < f(\textbf{b}(\alpha, W_\alpha.\hat{y}))$}{
%             $W_\alpha.\hat{y} \gets W_\alpha.y_\beta$
%         }
%     }
%     \For{each particle $\beta \in \{1 ... swarm\_size\}$}{
%         $W_\alpha.\hat{y}'_i \gets best(W_\alpha.y_{\beta-1}, W_\alpha.y_\beta, W_\alpha.y_{\beta+1})$
%     }
%     \If{$f(\textbf{b}(\alpha, W_\alpha.\hat{y})) < f(global\_best)$}{
%         $\alpha$th section of global\_best $\gets W_\alpha.\hat{y}$
%     }
%   }
%   \For{each swarm $\alpha \in \{1 ... m\}$}{
%     \For{each particle $\beta \in \{1 ... swarm\_size\}$}{
%         $W_\alpha.x_\beta \gets position\_update(W_\alpha.x_\beta)$
%     }
%   }
%   $loop\_counter += 1$
%  }
%  \caption{CCPSO2 algorithm} \label{alg:CCPSO2}
% \end{algorithm}

\subsection{Structure Stream} \label{sec:StructureStream}

Here, we aim to learn a sequential learner by using the spatial information of the extracted facial landmarks. Specifically, the input embedding of this sequential learner is determined by the matrix $U$. As shown in Figure~\ref{fig:pipeline}, the inputs to the structure stream are the matrix $\lambda$ which contains the spatial information of the facial landmarks, together with the $U$ matrix. These two matrices are used to form the input embedding for the sequential learner, for example an LSTM with peephole connections~\cite{gers2000recurrent}. The input embedding $\zeta$ is generated by:
\begin{equation}\label{Eq:StructInput}
\zeta = \lambda \times U.
\end{equation}

An LSTM network consists of a number of cells, the output of which continuously evolves with regards to its memory content. Moreover, each cell has a common cell state which is necessary for keeping track of long-term dependencies within the LSTM network. Two gates control the follow of information in the common cell state: input and forget gates, which make the LSTM cell decide when to forget or update the cell's state. The output of an LSTM cell is its hidden state, which is also controlled by the common cell state. We input $\zeta$ to the LSTM network and the aggregation of hidden states makes the final output of the LSTM network. 

% The intuition behind defining $\zeta$ as Equation~\ref{Eq:StructInput} is that basically, every column vector inside the matrix $U=[u_{\alpha} u_{\beta} \cdots u_{\gamma}]$ is calculating the linear combinations of the column vectors of matrix $\lambda$. Considering the way we defined the column vectors of matrix $U$, column vectors of the matrix $\zeta$ will be one of the $\lambda_1$ to $\lambda_{n}$. In other words, 
% The input embedding ($\zeta$) of the sequential learner is the extracted facial landmarks in the order which is determined by the matrix $U$ (see the input embedding in the Figure~\ref{fig:TreeTraversal}). 
% Please note that as traversing the tree by using Preorder depth first order may visit some nodes more than once, see Figure~\ref{fig:TreeTraversal}, $\zeta$ may include repeated landmarks. In other words, the relevant landmarks appear next to each other in the matrix $\zeta$. In the literature, it is established that the if the relevant landmarks appear next to each other, it may help the sequential learner to detect the dependencies more more efficiently~\cite{liu2017skeleton}.

Next, we incorporate an attention mechanism ~\cite{rocktaschel2015reasoning} over the output of our sequential learner to focus on the salient parts of the embedding. To do so, each hidden state vector of the sequential learner cell ($h_i$) is multiplied by a learnable weight $\delta_i$. $\delta_i$ controls the amount of attention each $h_i$ receives and can be expressed as:
\begin{equation}\label{Eq:attentionWeights}
\delta_i = \frac{e^{o_i}}{\sum_{k=1}^n e^{o_k}},
\end{equation}
where $n$ illustrates the number of units in the sequential learner and $o_i$ can be calculated using the following equation:
\begin{equation}\label{Eq:attentionUi}
o_i = tanh(W_hh_i+b_h).
\end{equation}
Here, $W_h$ and $b_h$ denote the set of trainable weights and biases respectively. We calculate the final attentive output of the structure stream $H_{structure}$ using:
\begin{equation}\label{Eq:attentionFinal}
H_{structure} = \sum_i\delta_i h_i.
\end{equation}
% where, $H_{structure}$ is the structure stream's final embedding.

\subsection{Texture Stream} \label{sec:TextureStream}
In this stream, the texture information of the input images is used to learn a sequential learner. Like the structure stream, the matrix $U$ determines the input embedding of the sequential learner. Inputs to this stream are the input image $I$, the matrix $\lambda$, and the matrix $U$. To use the texture information of the input image $I$, we first create $a \times a$ patches around the center of each facial landmark (the optimal value of parameter $a$ is determined empirically in Section~\ref{sec:PatchSize}. The patches are then cropped and fed to a pre-trained encoder, such as ResNet-50 pre-trained on the VGG-Face2 dataset~\cite{cao2018vggface2}.
%We conduct many experiments to determine the best $a$, which will be presented in the experiments section (Section~\ref{sec:PatchSize}). 

Suppose that the encoder generates an embedding for the first patch. We call this embedding $\psi_1$. We stack the outputs of the encoder for all the patches in a single matrix denoted by $\Psi = [\psi_1 \psi_2 \cdots \psi_n]$. 
%where $\psi_i$ represents the obtained embedding for the patch formed around facial landmark $i$. 
By multiplying $\Psi$ and $U$, we compute the input embedding of the second sequential learner (recall, the first sequential learner was used in the structure stream). 
%We input this embedding to the sequential learner. 
%The hidden states of the sequential learner's cells comprise its final output.
Finally, we take the output of this sequential learner from its cells' hidden states and employ an attention mechanism on them similar to the structure stream.
% following the sequential learner (see equations~\ref{Eq:attentionWeights} through \ref{Eq:attentionFinal}).

\subsection{Fusion}
\begin{figure}
    \begin{center}
    \includegraphics[width=\columnwidth]{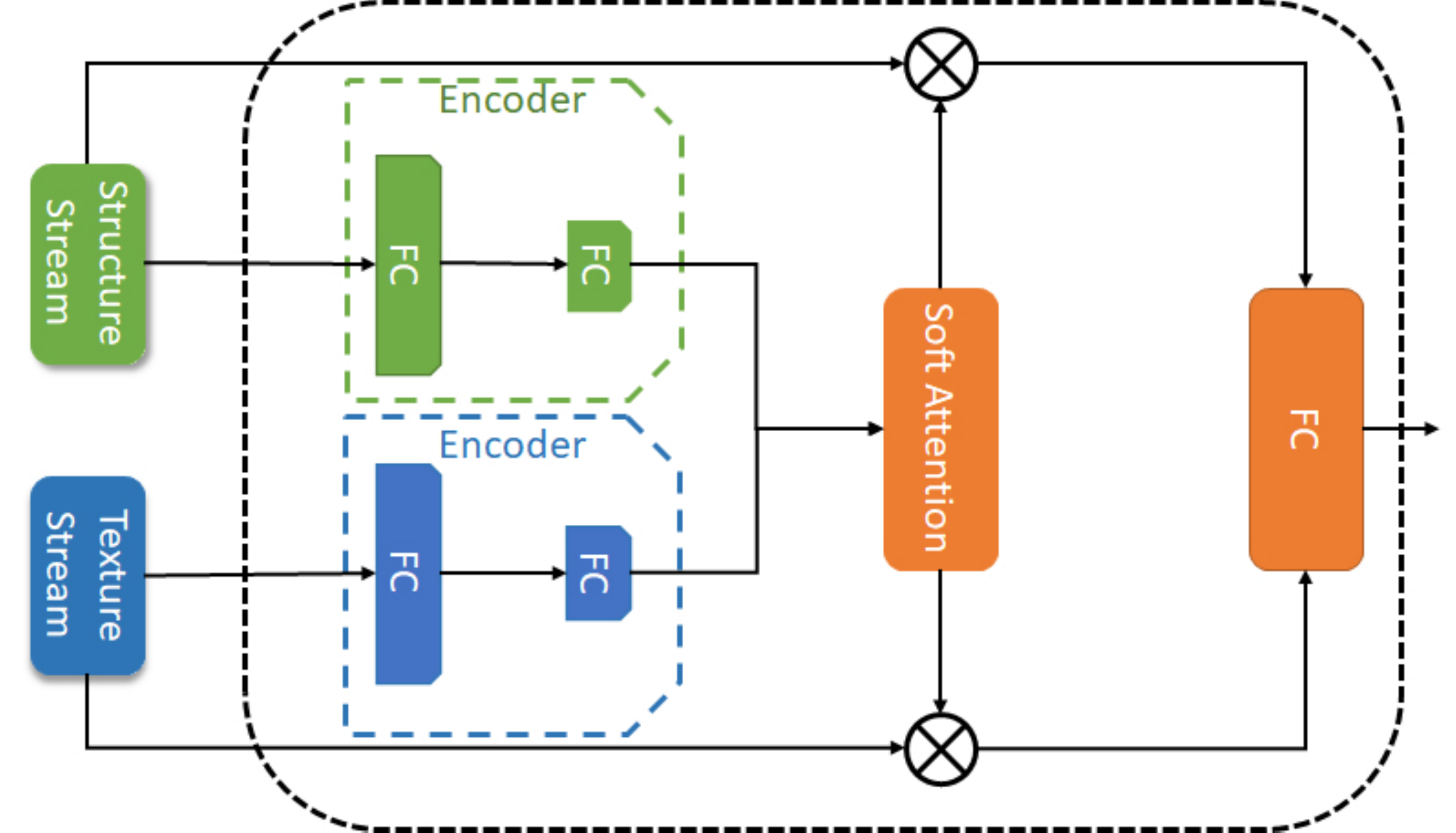}
    \end{center}
\caption{Illustration of the fusion strategy employed in FaceTopoNet.}
\label{fig:fusion}
\end{figure}

To combine the outputs of the two attention mechanisms at the end of each of the streams, we employ the fusion strategy proposed in~\cite{gu2018hybrid} and used in other prior work such as~\cite{zhang2020rfnet}. 
% to fuse the structure and texture streams outputs. 
This fusion strategy was initially created with the goal of preserving the original properties of its input embeddings, and has obtained state-of-the-art results in many areas~\cite{gu2018hybrid}. We add two dense layers for the output of structure and texture streams. These dense layers act like an encoder to generate stream-specific embeddings. We then employ soft attention to learn the weight $\eta$ using the following equation:
\begin{equation}\label{Eq:fusion1}
\eta=softmax(tanh(W_f[T^*,S^*]+b_f)),
\end{equation}
where $W_f$ and $b_f$ respectively denote the trainable weights and biases, and $T^*$ and $S^*$ respectively denote the stream-specific embeddings from texture and structure streams. Finally, for learning the optimal combination of the two stream-specific embeddings, we utilize a dense layer as:
\begin{equation}\label{Eq:fusion2}
y=tanh(W_y[(1+\eta_T)H_{texture}, (1+\eta_S)H_{structure}]+b_y),
\end{equation}
where $y$ denotes the final embedding and $W_y$ and $b_y$ represent learnable weights and biases. Finally, $y$ is input to a softmax layer for determining the expression class probability. Figure~\ref{fig:fusion} summarizes the fusion strategy used in this work.

\subsection{Loss Function and Training} \label{sec:loss}
Considering each training epoch, a set of weights related to $K_n$ is generated by the CCPSO2 algorithm. These weights are used to form the $MST$, the traversal of which yields the matrix $U$. Next, we freeze the weights of $K_{n}$ and matrix $U$ and fully train the following texture and structure streams using ADAM optimizer~\cite{kingma2014adam}. When the training of the two streams is complete, we feed the final loss value to the CCPSO2 optimization algorithm as its loss function. Given the new loss value, the optimization algorithm creates a set of new weights for the complete graph. This process repeats until convergence.

\begin{figure}
    \begin{center}
    \includegraphics[scale=0.32]{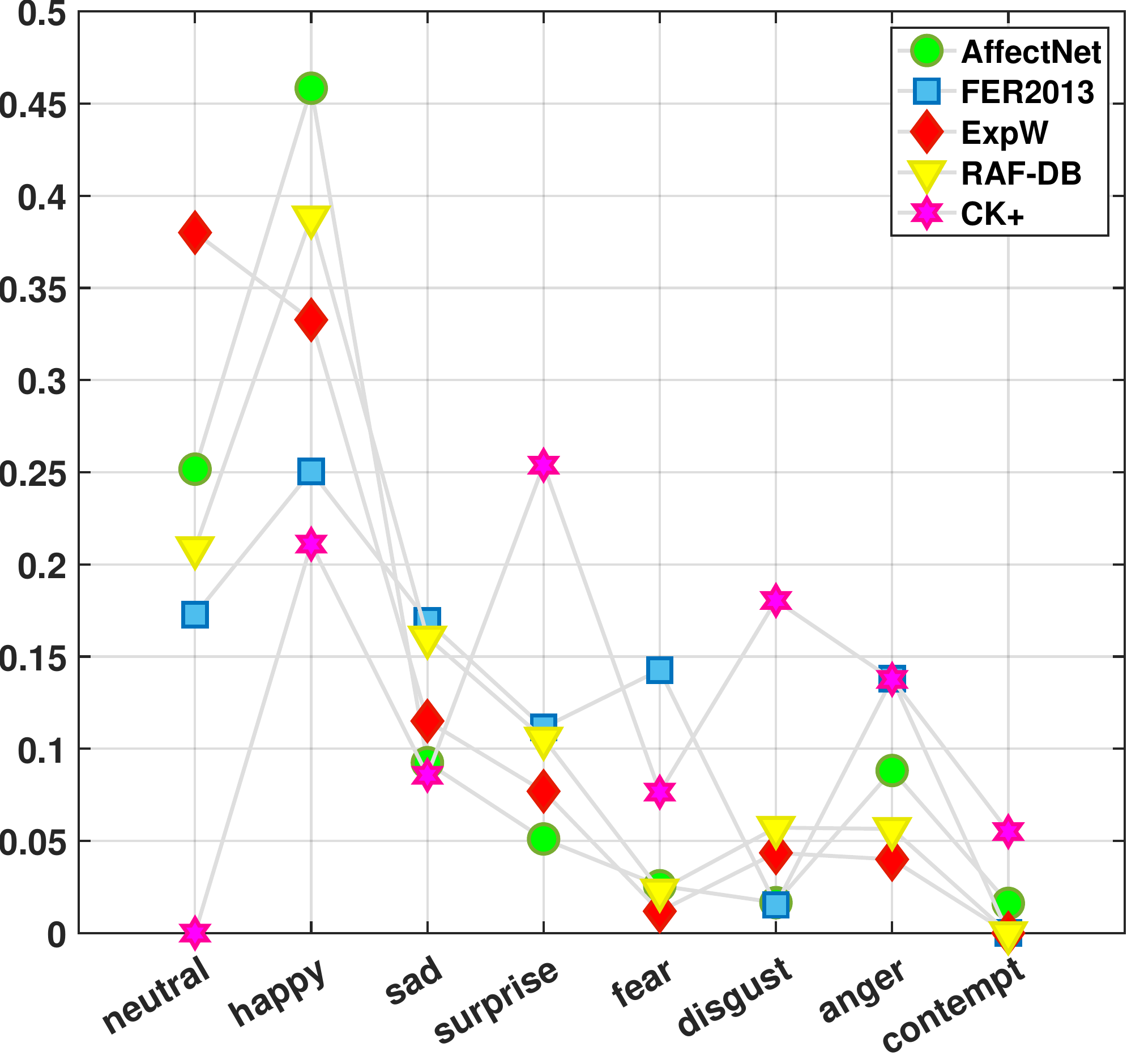}
    \end{center}
\caption{Distribution of expressions in AffectNet, FER2013, RAF-DB, and ExpW datasets.}
\label{fig:PMF}
\end{figure}

We define the total loss ($\mathcal{L}_{total}$) for both CCPSO2 and ADAM optimizers in our end-to-end method as:
\begin{equation}\label{Eq:loss}
\mathcal{L}_{total}=\frac{1}{3}(L_1+L_2+L_3),
\end{equation}
% \begin{multline}\label{Eq:loss}
% J(w)=\frac{1}{3m}\sum\limits_{i=1}^m L_1(\hat{y}^{(i)}, y^i ) + \frac{1}{3m}\sum\limits_{i=1}^m L_2(\hat{y}^{(i)}, y^i ) +\\ \frac{1}{3m}\sum\limits_{i=1}^m L_3(\hat{y}^{(i)}, y^i), 
% \end{multline}
% where \textit{m} represents the training set size, $L_1(x)$ denotes the fusion loss function, and $L_2(x)$ and $L_3(x)$ represent the loss functions of the structure and texture streams respectively.
where $L_1$ denotes the fusion loss, and $L_2$ and $L_3$ represent the losses for the structure and texture streams respectively.
Generally, FER datasets are highly skewed towards a specific expression, usually happy.  To demonstrate this, we depict the probability mass function (pmf) of the datasets used in this study (AffectNet, FER2013, RAF-DB, ExpW, and CK+) in Figure~\ref{fig:PMF}. This figure clearly demonstrates that the distributions are highly skewed toward Happy class. This issue makes it hard for FER algorithms to perform well on minority classes~\cite{he2009learning}. To address this issue, we adopt focal loss~\cite{lin2017focal} for all three loss terms in Equation~\ref{Eq:loss}.
% in order to calculate the loss between $\hat{y}^{(i)}$ and ${y}^{(i)}$. 
Therefore, the loss terms are defined as:
\begin{equation}\label{Eq:focal}
L_i = -\alpha_b(1-p_b)^\gamma log(p_b)~~~\text{for}~i=1, 2, 3 ,
\end{equation}
where $\alpha_b$ is a balancing factor and $p_b$ is defined by:
\begin{equation}\label{Eq:pb}
p_b= \begin{cases} 
      p & y=1\\
      1-p & otherwise ,
      \end{cases}
\end{equation}
where $y$ is the ground truth and $p$ is the probability calculated by the model. In Equation~\ref{Eq:focal}, $\gamma$ denotes the focusing parameter. Empirically, we set $\gamma$ and $\alpha_b$ to 2 and 0.25, respectively. Our use of focal loss allows the model to become less focused on the majority class by down-weighting the samples which are well-classified as shown in the equation.

\subsection{Implementation Details}
We implement the proposed architecture using TensorFlow~\cite{abadi2016tensorflow} and train it using two Nvidia RTX 2080 Ti GPUs. For ADAM optimizer, we set learning rate, first-momentum decay, and second-momentum-decay as 0.001, 0.9, and 0.99 respectively. For all the experiments, the face images are cropped and resized to 60$\times$60$\times$3 pixels. Additionally, for data augmentation purpose, each image is horizontally flipped with the probability of 0.25. Similar to \cite{hasani2020breg, kollias2020deep, vielzeuf2017temporal}, the reported results are achieved on the validation set of the respective datasets. However, for CK+ dataset, we use leave-one-subject-out cross validation~\cite{lucey2010extended}

Considering CCPSO2 algorithm, the number of iterations is set to 40, as convergence occurs at or before this iteration, after which no changes are observed in the generated trees. Additionally, for random grouping of the variables we use the set $\{1, 5, 7\}$ and the population size for each swarm is set to 10. As CCPSO2 algorithm is a stochastic method, we run it 30 times. However, the converged tree did not change, indicating the consistency of the approach.

% *****************************************************************************************************************************

% \usepackage{multirow}

% \usepackage{color}
% \usepackage{multirow}

\begin{table*}
\centering
\caption{Obtained recognition rates for different encoders and sequential learners.}
\label{table:seq-enc}
\begin{tabular}{l|l|l|l|l|l|l} 
\hline
\multirow{2}{*}{\textbf{Dataset}}      & \multirow{2}{*}{\textbf{Encoder}} & \multirow{2}{*}{\textbf{Pretrained on}} & \multicolumn{4}{c}{\textbf{Sequential Learner}}                                                                        \\ 
\cline{4-7}
                                       &                                   &                                         & \textbf{LSTM}                    & \textbf{Bi-LSTM}       & \textbf{GRU}                     & \textbf{Bi-GRU}         \\ 
\hline
\multirow{4}{*}{AffectNet}             & ResNet-50                         & VGG-Face2                               & \textbf{70.02}                   & 69.81                  & 69.91                            & 69.53                   \\
                                       & ResNet-50                         & ImageNet                                & 69.37                            & 68.90                  & 69.23                            & 69.17                   \\
                                       & VGG-16                            & ImageNet                                & 68.21                            & 68.28                  & 68.72                            & 68.42                   \\
                                       & BreG-NeXt-50~                     & VGG-Face2                               & 69.43                            & 69.15                  & 69.57                            & 69.43                   \\ 
\hline
\multirow{4}{*}{FER2013}               & ResNet-50                         & VGG-Face2                               & \textbf{72.66}                   & 72.51                  & 72.57                            & 72.43                   \\
                                       & ResNet-50                         & ImageNet                                & 71.26                            & 71.26                  & 71.31                            & 71.23                   \\
                                       & VGG-16                            & ImageNet                                & 71.21                            & 71.19                  & 71.21                            & 71.14                   \\
                                       & BreG-NeXt-50                      & VGG-Face2                               & 72.48                            & 72.37                  & 72.46                            & 72.31                   \\ 
\hline
\multirow{3}{*}{ExpW}                  & ResNet-50                         & VGG-Face2                               & \textbf{71.91}                   & 71.83                  & 71.83                            & 71.78                   \\
                                       & ResNet-50                         & ImageNet                                & 70.74                            & 70.66                  & 70.75                            & 70.75                   \\
                                       & VGG-16                            & ImageNet                                & 70.49                            & 70.49                  & 70.54                            & 70.51                   \\ 
\hline
\multirow{3}{*}{RAF-DB}                & ResNet-50                         & VGG-Face2                               & 87.80                            & 87.78                  & \textbf{87.81}                   & 87.74                   \\
                                       & ResNet-50                         & ImageNet                                & 86.93                            & 86.95                  & 87.02                            & 86.87                   \\
                                       & VGG-16                            & ImageNet                                & 86.82                            & 86.77                  & 86.94                            & 86.74                   \\ 
\hline
\multirow{3}{*}{CK+} & ResNet-50       & VGG-Face2             & \textbf{99.19} & 98.9 & \textbf{99.19} & 97.9  \\
                                       & ResNet-50       & ImageNet              & 98.4           & 97.8 & 97.3           & 97.1  \\
                                       & VGG-16          & ImageNet              & 96.3           & 95.9 & 96.1           & 96.3  \\
\hline
\end{tabular}
\end{table*}

\section{Experiments} \label{sec:experiments}
We use four well-known public in-the-wild FER datasets to evaluate our method. In this section, we first introduce these datasets, followed by the details of our experiments, and obtained results. 

\subsection{Datasets}
\textbf{AffectNet \cite{Affectnet}:} Introduced in 2017, AffectNet includes more than 1 million face images. All the images are collected from the Internet using three different search engines. Within the dataset, about 450,000 images have been annotated manually while the remaining have been automatically annotated, i.e. using deep neural networks. Additionally, this dataset provides the ground truth for both the categorical model of affect (happy, sad, surprise, disgust, fear, anger, contempt, and neutral) and the dimensional model of affect, i.e. arousal and valence values for each image. The validation set of AffectNet includes 500 images for each expression class.

\textbf{FER2013 \cite{FER2013}:} Introduced in 2013, FER2013 comprises 28,709 and 7,178 images respectively for training and validation. All the images have the resolution of 48 $\times$ 48 pixels. Moreover, this dataset labels each image in one of the following 6 expression classes: happy, sad, surprise, disgust, fear, anger, and neutral.

\textbf{RAF-DB \cite{li2017reliable}}: The real-world affective face database (RAF-DB) provides 29,672 face images with their associated expression labels. Crowdsourcing has been used to assign each face image to one of the 7 basic and 11 compound expression classes. Additionally, the dataset has been split into train and validation sets to provide a standard benchmark to the community.

\textbf{ExpW \cite{SOCIALRELATION_2017}}: This dataset is collected from the Internet by using the Google search engine and offers 91,793 face images. All the images have been labeled manually as one of the 7 basic expression classes.

\textbf{CK+ \cite{lucey2010extended}}: Comprising 327 videos, the extended Cohn-Kanade (CK+) dataset is one of the well-known precursors in the FER datasets. Each video sequence demonstrates a change from neutral to the peak expression. Moreover, CK+ provides the expression labels as being one of the six basic expressions plus contempt expression. For evaluation purpose, we select three peak expression frames within each expression sequence to build train and validation sets.

\begin{table*}
\centering

\setlength
\tabcolsep{4pt}
\caption{Results of FaceTopoNet on AffectNet dataset. `Ne.', `Ha.', `Sa.' `Su.', `Fe.', `Di.', `An.', and `Co.', denote neutral, happy, sad, surprise, fear, disgust, anger, and contempt respectively.}
\label{table:resultsAffect}
\begin{tabular}{l|l|l|l|l|l|l|l|l|l|l|l|l|l|l|l|l|l|l} 
\hline
\multirow{2}{*}{\textbf{Authors}} & \multirow{2}{*}{\textbf{Method}} & \multirow{2}{*}{\textbf{RR}} & \multicolumn{8}{l|}{\textbf{Precision}}                                                                                                                                                                                                                                                                               & \multicolumn{8}{l}{\textbf{Recall}}                                                                                                                                                                                                                                                                                    \\ 
\cline{4-19}
                                                                            &                                  &                              & Ne.                                  & Ha.                                  & Sa.                                  & Su.                                  & Fe.                                  & Di.                                  & An.                                  & Co.                                  & Ne.                                  & Ha.                                  & Sa.                                  & Su.                                  & Fe.                                  & Di.                                  & An.                                  & Co.                                   \\ 
\hline
Hewitt  et al.~\cite{hewitt2018cnn}                                                             & CNN                              & 58                           & 45                                   & 76                                   & 59                                   & 55                                   & 60                                   & 55                                   & 57                                   & 59                                   & 49                                   & 72                                   & 56                                   & 60                                   & \textbf{64}                          & 63                                   & 43                                   & 56                                    \\
Hua et al.~\cite{hua2019hero}                                                                 & Ensemble                         & 62.11                        & 58                                   & 72                                   & 64                                   & 62                                   & 57                                   & 59                                   & 62                                   & -                                    & \textbf{59}                          & 70                                   & \textbf{67}                          & 66                                   & 61                                   & 59                                   & 57                                   & {\cellcolor[rgb]{0.753,0.753,0.753}}  \\
Hasani et al.~\cite{hasani2020breg}                                                              & BreG-NeXt32                      & 66.74                        & {\cellcolor[rgb]{0.753,0.753,0.753}} & {\cellcolor[rgb]{0.753,0.753,0.753}} & {\cellcolor[rgb]{0.753,0.753,0.753}} & {\cellcolor[rgb]{0.753,0.753,0.753}} & {\cellcolor[rgb]{0.753,0.753,0.753}} & {\cellcolor[rgb]{0.753,0.753,0.753}} & {\cellcolor[rgb]{0.753,0.753,0.753}} & {\cellcolor[rgb]{0.753,0.753,0.753}} & {\cellcolor[rgb]{0.753,0.753,0.753}} & {\cellcolor[rgb]{0.753,0.753,0.753}} & {\cellcolor[rgb]{0.753,0.753,0.753}} & {\cellcolor[rgb]{0.753,0.753,0.753}} & {\cellcolor[rgb]{0.753,0.753,0.753}} & {\cellcolor[rgb]{0.753,0.753,0.753}} & {\cellcolor[rgb]{0.753,0.753,0.753}} & {\cellcolor[rgb]{0.753,0.753,0.753}}  \\
Hasani et al.~\cite{hasani2020breg}                                                              & BreG-NeXt50                      & 68.50                        & \textbf{72}                          & \textbf{78}                          & 62                                   & \textbf{66}                          & 67                                   & 62                                   & 63                                   & 67                                   & 53                                   & 89                                   & 66                                   & 74                                   & 63                                   & 77                                   & 61                                   & 58                                    \\
Chen et al.~\cite{chen2021residual}                                                                 & Residual multi-task learning     & 61.98                        & {\cellcolor[rgb]{0.753,0.753,0.753}}                                     & {\cellcolor[rgb]{0.753,0.753,0.753}}                                     &{\cellcolor[rgb]{0.753,0.753,0.753}}                                      &{\cellcolor[rgb]{0.753,0.753,0.753}}                                      &{\cellcolor[rgb]{0.753,0.753,0.753}}                                      &{\cellcolor[rgb]{0.753,0.753,0.753}}                                      & {\cellcolor[rgb]{0.753,0.753,0.753}}                                     & {\cellcolor[rgb]{0.753,0.753,0.753}}                                     &{\cellcolor[rgb]{0.753,0.753,0.753}}                                      &{\cellcolor[rgb]{0.753,0.753,0.753}}                                      &{\cellcolor[rgb]{0.753,0.753,0.753}}                                      &{\cellcolor[rgb]{0.753,0.753,0.753}}                                      &{\cellcolor[rgb]{0.753,0.753,0.753}}                                      & {\cellcolor[rgb]{0.753,0.753,0.753}}                                     &{\cellcolor[rgb]{0.753,0.753,0.753}}                                      &{\cellcolor[rgb]{0.753,0.753,0.753}}                                       \\
Zhang et al.~\cite{zhang2021learning}                                                                & Pseudo-siamese
structure~ ~      & 63.70                        &{\cellcolor[rgb]{0.753,0.753,0.753}}                                      & {\cellcolor[rgb]{0.753,0.753,0.753}}                                     &{\cellcolor[rgb]{0.753,0.753,0.753}}                                      &{\cellcolor[rgb]{0.753,0.753,0.753}}                                      &{\cellcolor[rgb]{0.753,0.753,0.753}}                                      &{\cellcolor[rgb]{0.753,0.753,0.753}}                                      &{\cellcolor[rgb]{0.753,0.753,0.753}}                                      &{\cellcolor[rgb]{0.753,0.753,0.753}}                                      &{\cellcolor[rgb]{0.753,0.753,0.753}}                                      &{\cellcolor[rgb]{0.753,0.753,0.753}}                                      & {\cellcolor[rgb]{0.753,0.753,0.753}}                                     & {\cellcolor[rgb]{0.753,0.753,0.753}}                                     &{\cellcolor[rgb]{0.753,0.753,0.753}}                                      &{\cellcolor[rgb]{0.753,0.753,0.753}}                                      & {\cellcolor[rgb]{0.753,0.753,0.753}}                                     & {\cellcolor[rgb]{0.753,0.753,0.753}}                                      \\
Farzaneh et al.\cite{farzaneh2021facial}                                                             & Deep attentive center loss       & 65.20                        &{\cellcolor[rgb]{0.753,0.753,0.753}}                                      &{\cellcolor[rgb]{0.753,0.753,0.753}}                                      &{\cellcolor[rgb]{0.753,0.753,0.753}}                                      &{\cellcolor[rgb]{0.753,0.753,0.753}}                                      &{\cellcolor[rgb]{0.753,0.753,0.753}}                                      &{\cellcolor[rgb]{0.753,0.753,0.753}}                                      &{\cellcolor[rgb]{0.753,0.753,0.753}}                                      &{\cellcolor[rgb]{0.753,0.753,0.753}}                                      &{\cellcolor[rgb]{0.753,0.753,0.753}}                                      &{\cellcolor[rgb]{0.753,0.753,0.753}}                                      &{\cellcolor[rgb]{0.753,0.753,0.753}}                                      &{\cellcolor[rgb]{0.753,0.753,0.753}}                                      &{\cellcolor[rgb]{0.753,0.753,0.753}}                                      &{\cellcolor[rgb]{0.753,0.753,0.753}}                                      &{\cellcolor[rgb]{0.753,0.753,0.753}}                                      &{\cellcolor[rgb]{0.753,0.753,0.753}}                                       \\
\textbf{Ours}                                                               & FaceTopoNet                      & \textbf{70.02}               & \textbf{72}                          & 63                                   & \textbf{75}                          & 65                                   & \textbf{68}                          & \textbf{72}                          & \textbf{74}                          & \textbf{77}                          & 38                                   & \textbf{98}                          & 66                                   & \textbf{78}                          & 63                                   & \textbf{88}                          & \textbf{68}                          & \textbf{59}                           \\
\hline
\end{tabular}

\end{table*}

\begin{table*}
\centering

\setlength
\tabcolsep{4pt}
\caption{Results of FaceTopoNet on FER2013 dataset.}
\label{table:resultsFER}
\begin{tabular}{l|l|l|l|l|l|l|l|l|l|l|l|l|l|l|l|l} 
\hline
\multirow{2}{*}{\begin{tabular}[c]{@{}l@{}}\\\textbf{Authors}\end{tabular}} & \multirow{2}{*}{\textbf{Method}} & \multirow{2}{*}{\textbf{RR}} & \multicolumn{7}{l|}{\textbf{Precision}}                                                                                                                                                                                                                                        & \multicolumn{7}{l}{\textbf{Recall}}                                                                                                                                                                                                                                             \\ 
\cline{4-17}
                                                                            &                                  &                              & Ne.                                  & Ha.                                  & Sa.                                  & Su.                                  & Fe.                                  & Di.                                  & An.                                  & Ne.                                  & Ha.                                  & Sa.                                  & Su.                                  & Fe.                                  & Di.                                  & An.                                   \\ 
\hline
Vielzeuf et al.~\cite{vielzeuf2017temporal}                                                            & CNN                              & 71.2                         & {\cellcolor[rgb]{0.753,0.753,0.753}} & {\cellcolor[rgb]{0.753,0.753,0.753}} & {\cellcolor[rgb]{0.753,0.753,0.753}} & {\cellcolor[rgb]{0.753,0.753,0.753}} & {\cellcolor[rgb]{0.753,0.753,0.753}} & {\cellcolor[rgb]{0.753,0.753,0.753}} & {\cellcolor[rgb]{0.753,0.753,0.753}} & {\cellcolor[rgb]{0.753,0.753,0.753}} & {\cellcolor[rgb]{0.753,0.753,0.753}} & {\cellcolor[rgb]{0.753,0.753,0.753}} & {\cellcolor[rgb]{0.753,0.753,0.753}} & {\cellcolor[rgb]{0.753,0.753,0.753}} & {\cellcolor[rgb]{0.753,0.753,0.753}} & {\cellcolor[rgb]{0.753,0.753,0.753}}  \\
Hasani et al.~\cite{hasani2020breg}                                                              & BreG-NeXt32                      & 69.11                        & {\cellcolor[rgb]{0.753,0.753,0.753}} & {\cellcolor[rgb]{0.753,0.753,0.753}} & {\cellcolor[rgb]{0.753,0.753,0.753}} & {\cellcolor[rgb]{0.753,0.753,0.753}} & {\cellcolor[rgb]{0.753,0.753,0.753}} & {\cellcolor[rgb]{0.753,0.753,0.753}} & {\cellcolor[rgb]{0.753,0.753,0.753}} & {\cellcolor[rgb]{0.753,0.753,0.753}} & {\cellcolor[rgb]{0.753,0.753,0.753}} & {\cellcolor[rgb]{0.753,0.753,0.753}} & {\cellcolor[rgb]{0.753,0.753,0.753}} & {\cellcolor[rgb]{0.753,0.753,0.753}} & {\cellcolor[rgb]{0.753,0.753,0.753}} & {\cellcolor[rgb]{0.753,0.753,0.753}}  \\
Hasani et al.~\cite{hasani2020breg}                                                              & BreG-NeXt50                      & 71.53                        & 69                                   & 88                          & 59                                   & 78                          & 52                                   & 62                          & 60                                   & 69                          & 90                                   & 62                                   & 80                                   & 46                                   & 12                                   & 65                                    \\
Shi et al.~\cite{shi2021facial}                                                                  & cross-connected CNN              & 71.52                        & {\cellcolor[rgb]{0.753,0.753,0.753}}                                     &{\cellcolor[rgb]{0.753,0.753,0.753}}                                      &{\cellcolor[rgb]{0.753,0.753,0.753}}                                      &{\cellcolor[rgb]{0.753,0.753,0.753}}                                      &{\cellcolor[rgb]{0.753,0.753,0.753}}                                      &{\cellcolor[rgb]{0.753,0.753,0.753}}                                      &{\cellcolor[rgb]{0.753,0.753,0.753}}                                      &{\cellcolor[rgb]{0.753,0.753,0.753}}                                      &{\cellcolor[rgb]{0.753,0.753,0.753}}                                      &{\cellcolor[rgb]{0.753,0.753,0.753}}                                      &{\cellcolor[rgb]{0.753,0.753,0.753}}                                      &{\cellcolor[rgb]{0.753,0.753,0.753}}                                      &{\cellcolor[rgb]{0.753,0.753,0.753}}                                      &{\cellcolor[rgb]{0.753,0.753,0.753}}                                       \\
Pourramezan et al.~\cite{fard2022ad}                                                          & Deep metric learning             & 72.03                        &63                                      &\textbf{90}                                      &61                                      &\textbf{80}                                      & 61                                     & \textbf{75}                                     &63                                      &\textbf{73}                                      &88                                      &56                                      &\textbf{83}                                      &\textbf{59}                                      & \textbf{77}                                     &62                                       \\
\textbf{Ours}                                                               & FaceTopoNet                      & \textbf{72.66}               & \textbf{71}                          & 63                                   & \textbf{63}                          & 70                                   & \textbf{63}                          & 42                                   & \textbf{66}                          & 64                                   & \textbf{93}                          & \textbf{64}                          & 82                          & 54                          & 20                          & \textbf{73}                           \\
\hline
\end{tabular}

\end{table*}

% \subsection{Encoder-Sequential Learner Experiment}
\subsection{Encoders and Sequential Learners}
First we aim to identify the best encoder and sequential learner to be used in FaceTopoNet. 
% For choosing the best encoder and sequential learner to be used in the FaceTopoNet (see figure~\ref{fig:pipeline}), 
Accordingly, we consider several candidates for each sub-module. For the sequential learner we test LSTM, bidirectional LSTM (Bi-LSTM)~\cite{graves2013speech}, gated recurrent unit (GRU)~\cite{cho2014learning}, and bidirectional GRU (Bi-GRU)~\cite{cho2014learning}, while for the encoder, we explore ResNet-50 pre-trained on VGG-Face2, ResNet-50 pre-trained on ImageNet~\cite{deng2009imagenet}, VGG-16 pre-trained on ImageNet, and BReG-NeXt-50 pretrained on VGG-Face2 for encoder. These sub-modules have been selected for experimentation given their strong performance in prior work~\cite{hasani2020breg, rajan2020novel}. We report the FER results for each dataset in Table~\ref{table:seq-enc}. Given the results, the best combination for 4 of the datasets is achieved with a ResNet-50 pretrained on VGG-Face2 and LSTM. The only exception is on the RAF-DB dataset in which GRU exhibits slightly (marginal) better performance than the LSTM. Additionally, regarding CK+, GRU-ResNet-50 combination delivers the same accuracy as the LSTM-ResNet-50 combination. Hence, for the sake of consistency in our experiments, we use a ResNet-50 and LSTM in FaceTopoNet. 
% in the final version of the FaceTopoNet, we choose the sequential learner and the encoder as being LSTM and ResNet-50. 
We observe from the table that pre-training on VGG-Face2 obtains better results compared to ImageNet, which is  
% From the Table~\ref{table:seq-enc}, it is evident that in all experiments, ResNet-50 pre-trained on VGG-Face2 achieved higher RRs than the same ResNet-50 pre-trained on ImageNet. This behavior is 
predictable given that VGG-Face2 only includes face images with different poses and illuminations as opposed to ImageNet which contains various image classes of different objects. This can help the encoder to extract embeddings that are discriminative and important in the context of human faces. 
% In addition, it is clear that generally LSTMs and GRUs outperform their \textit{bidirectional} versions.
Moreover, we observe that the added depth and residual connections of ResNet-50 results in better performance compared to VGG-16.%, which is indicating that the skip connections are effective in FER and can boost the network's performance~\cite{hasani2020breg}.

\subsection{Performance}
We report the RR values for FaceTopoNet along with other state-of-the-art benchmarks on all 5 datasets in Tables ~\ref{table:resultsAffect} through \ref{table:resultsCK}. Considering AffectNet (Table~\ref{table:resultsAffect}), FaceTopoNet outperforms the state-of-the-art FER methods BreG-NeXt-50 and BreGNeXt-32 by 1.52\% and 3.28\%, respectively. 
% Please note that both BReG-NeXt-50 and BReG-NeXt-32 adopted the same focal loss as their loss function. 
With regards to FER2013 dataset (Table~\ref{table:resultsFER}), FaceTopoNet shows an improvement of 1.13\% over the state-of-the-art, BreG-NeXt-50. The results for ExpW are presented in Table~\ref{table:resultsExpW}. In this table, we can observe that FaceTopoNet has the highest recognition rate (71.91\%) compared to other methods. This is followed by~\cite{lian2020expression}, which also uses part-based methods, dividing the face into different sections, to perform FER. The results for RAF-DB are presented in Table~\ref{table:resultsRAF-DB}. As it is evident from this table, our method achieves the second-best recognition rate, being marginally below~\cite{li2021adaptively} by only 0.27\%. A reason for this could be the use of a new loss term by~\cite{li2021adaptively}, which is equipped to handle class imbalances, which is a common issue in FER datasets, by monitoring the last layers of a CNN. Finally, regarding CK+ dataset, FaceTopoNet achieves the second best recognition rate. The current state-of-the-art~\cite{ruan2021feature} offers the recognition rate of 99.54\%, while FaceTopoNet achieves slightly lower recognition rate, 99.19\%.
% \cite{li2021adaptively} outperforms FaceTopoNet by a small margin of 0.27\%.%, and the FaceTopoNet gives the second highest RR.

As some of the state-of-the-art methods additionally reported precision and recall values for both AffectNet and FER2013 datasets, we also consider these measures for each emotion class in Tables~\ref{table:resultsAffect} and~\ref{table:resultsFER}. 
% The precision and recall results have been presented for expression classes. 
As it is evident from Table~\ref{table:resultsAffect}, for most of the expression classes, FaceTopoNet outperforms other state-of-the-art methods in terms of precision and recall. More specifically, FaceTopoNet obtains the best precision and recall in disgust and contempt expressions, which are considered as more difficult classes due to their small numbers of training samples in the AffectNet dataset. Given FER2013 dataset, FaceTopoNet delivers the highest precision in most of the expression classes - i.e. neutral, sad, fear, and angry - while acheiving the state-of-the-art recall in 3 expression classes.

\begin{figure}
    \begin{center}
     \includegraphics[width=\columnwidth]{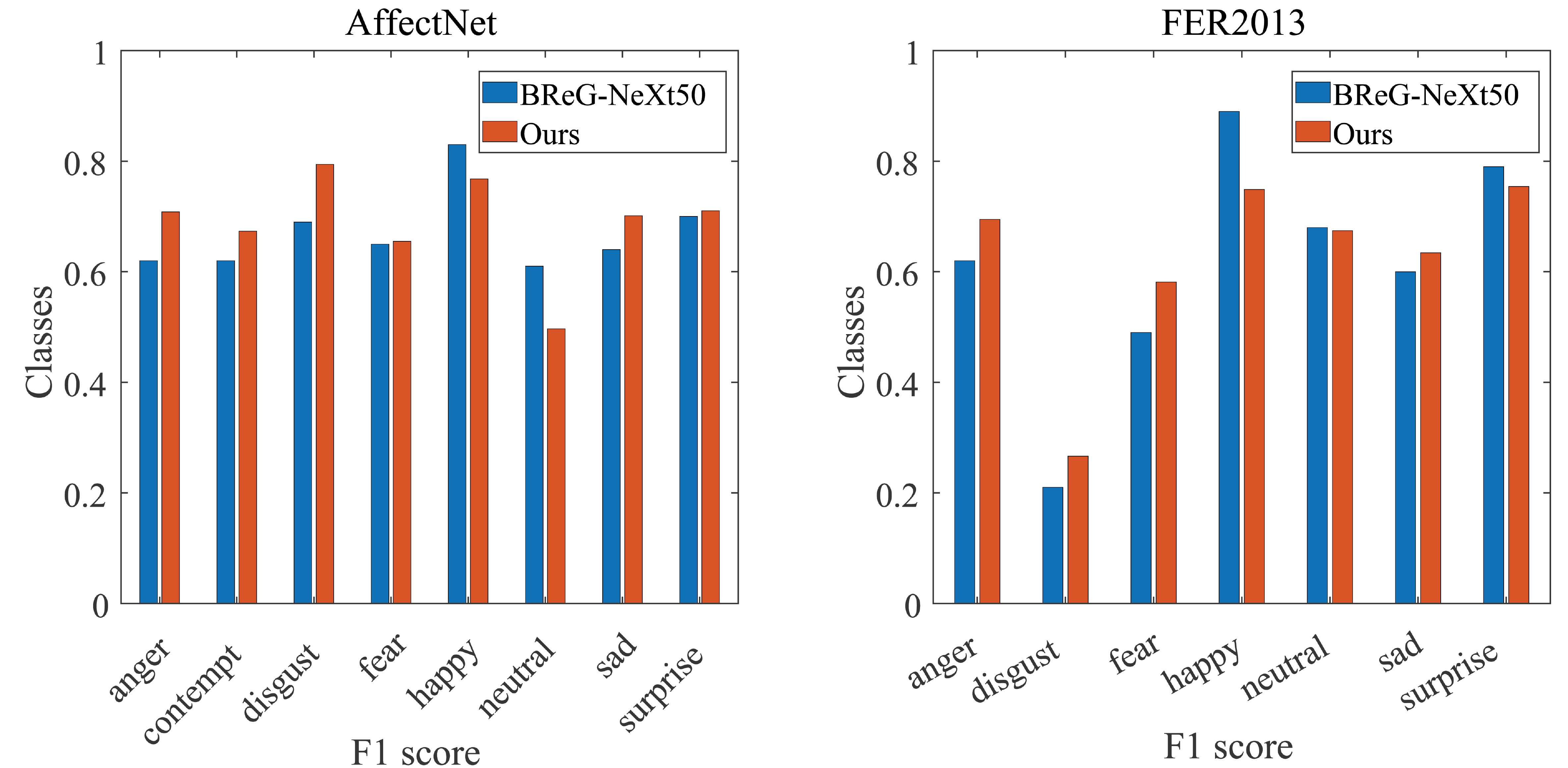}
    \end{center}
\caption{Achieved F1 scores of BreG-NeXt50 and FaceTopoNet.}
\label{fig:f1score}
\end{figure}

For a deeper comparison between our proposed model and one of the  best performing benchmarks in both AffectNet and FER2013 datasets, BreG-NeXt-50, F1 scores corresponding to the expression classes are presented in Figure~\ref{fig:f1score} using one-vs-all method~\cite{galar2011overview}. Considering AffectNet, our proposed method obtains higher F1 scores in 6 out of 8 expressions when compared to BreG-NeXt-50, while in FER2013 we outperform BreG-NeXt-50 in 4 of the 7 classes. %The only exceptions are neutral and happy expressions. As it is clear from the Figure~\ref{fig:PMF}, the pmf of expressions of AffectNet dataset is highly skewed toward neutral and happy. In spite of the mentioned imbalance, our proposed method performs better in the minority classes, recognizing of which is considered to be more difficult. 
% In terms of FER2013, FaceTopoNet shows higher F1 scores in most of the expressions with the exceptions of neutral, surprise, and happy expression classes.  

\begin{table}[t]
\centering

\caption{Results of FaceTopoNet on RAF-DB dataset}
\label{table:resultsRAF-DB}
\begin{tabular}{l|l|l} 
\hline
\textbf{Author} & \textbf{Method}          & \textbf{RR}  \\ 
\hline
Wang et al.~\cite{wang2021oaenet}   & pseudo-siamese network         & 86.50        \\
Wang et al.~\cite{wang2020region}      & Region attention         & 86.90        \\
Wang et al.~\cite{wang2020suppressing}      & Attention and relabeling & 87.03        \\
Farzaneh et al.\cite{farzaneh2021facial}  & CNN and attention        & 87.78        \\
Li et al.~\cite{li2021adaptively}        & Coarse-fine labeling                      & \textbf{88.07}        \\
Ours            & FaceTopoNet              & 87.80        \\
\hline
\end{tabular}
\end{table}

\begin{table}[t]
\centering

\caption{Results of FaceTopoNet on ExpW dataset.}
\label{table:resultsExpW}
\begin{tabular}{l|l|l} 
\hline
\textbf{Author}  & \textbf{Method}                  & \textbf{RR}  \\ 
\hline
Benamaraet et al.~\cite{benamara2021real} & Ensemble & 71.82        \\
Xie et al.\cite{GCN2}        & Adversarial graph representation & 68.50   \\
Lain et al.~\cite{lian2020expression}       & CNN-Prediction authenticity & 71.90        \\
Peng et al.~\cite{peng2022guided}                       & Domain adaptive FER       & 70.86\\
Ours             & FaceTopoNet                      & \textbf{71.91}        \\
\hline
\end{tabular}
\end{table}

\begin{table}[t]
\centering

\caption{Results of FaceTopoNet on CK+ dataset}
\label{table:resultsCK}
\begin{tabular}{l|l|l} 
\hline
\textbf{Author} & \textbf{Method}                                             & \textbf{RR}     \\ 
\hline
Ruan et al.~\cite{ruan2021feature}    & Feature decomposition and reconstruction                    & \textbf{99.54}  \\
Ruan et al.~\cite{ruan2020deep}    & removal of disturbing factors & 99.16           \\
Yang et al.~\cite{yang2018facial}    & De-expression
residue learning                              & 97.30           \\
Zeng et al.~\cite{zeng2018facialincon}     & Latent truth net                                            & 92.45           \\
Liu et al.~\cite{liu2017adaptive}      & combining the deep metric and softmax loss                  & 97.10           \\
Ours            & FaceTopoNet                                                 & 99.19           \\
\hline
\end{tabular}
\end{table}

\begin{figure*}
    \begin{center}
    \includegraphics[width=\textwidth]{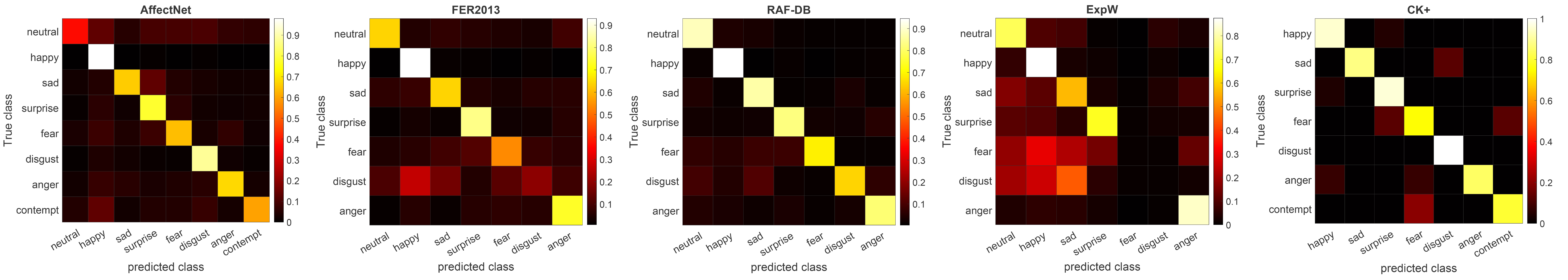}
    \end{center}
\caption{FaceTopoNet's Confusion Matrices.}
\label{fig:ConfMat}
\end{figure*}

We depict the obtained confusion matrices between the ground truth and the predicted expressions using our proposed method in Figure~\ref{fig:ConfMat}. For AffectNet, 3 cases, including contempt-happy, sad-surprise, and neutral-happy (the first and second elements are respectively the ground truth and the predicted expression) have the largest confusions. 
Confusion between contempt and happy is prevalent and expected in FER systems given the high semantic similarity of the two classes, and as evidenced by high confusion rates obtained from the perspective of human observers
% , because even human observers find it difficult to discriminate between the two expressions
~\cite{hasani2020breg}. 
Regarding FER2013, disgust-happy has the largest confusion. This is due to the fact that happy and disgust are respectively the majority and minority expression classes within the dataset causing a bias within the model (see the pmf for FER2013 in Figure~\ref{fig:PMF}). Despite the fact that we adopt focal loss to address the issue of the imbalanced training set, FaceTopoNet still identifies some of the disgust cases (minority class) as happy (majority class). Apart from disgust, FaceTopoNet performs well on the FER2013 dataset overall. 
% when distinguishing true classes. 
On the RAF-DB dataset, the largest confusions occur in the case of disgust-sad as well as disgust-neutral. The same issue can be seen in some of the prior works on RAF-DB~\cite{mavani2017facial}. As for the ExpW dataset, fear is confused with neutral, happy, and sad expressions. This can be justified as fear only constitutes 1.1\% of the ExpW dataset.
Finally, considering CK+ dataset, happy and disgust expressions have the highest accuracy. Additionally, the largest confusions occur in contempt-fear and sad-disgust cases. Among the expressions within the CK+, disgust expression shows the lowest error, while fear shows the highest error.

%We depict some of the true and false classification samples obtained by FaceTopoNet along with their output probabilities in Figure~\ref{fig:samples}. We observe that in many cases, the misclassified samples achieve the second highest probability with respect to the ground truth, indicating a very low rank-2 error for our model. 
The key concept behind the good performance of our method is the ``ordering'' of the embeddings. In Figure~\ref{fig:banner}, we have demonstrated that the ordering of the embeddings can affect the performance of sequential learners. Thus, our model optimizes the ordering of the embeddings, such that it becomes easy for a sequential learner to model the dependencies which exists among these embeddings. In this way, the sequential learner delivers superior performance.

% \begin{figure*}[t]
%     \begin{center}
%     \includegraphics[width=\textwidth]{fig9-mergedVersion_v2.pdf}
%     \end{center}
% \caption{Prediction samples of FaceTopoNet on AffectNet and FER2013. Ground truth of each image is mentioned below each sample. The color \textcolor{green}{green} indicates correct classification, while the color \textcolor{red}{red} indicated misclassification.}
% \label{fig:samples}
% \end{figure*}

\subsection{Optimal Number of Extracted Facial Landmarks} \label{sec:NnumberFacialLandmarks}
To choose the optimal value of the parameter $n$ (number of the extracted facial landmarks), we increase it by 10 each time and report the performance. The resulting curve is shown in Figure 8 for both AffectNet and FER2013 datasets. Considering AffectNet dataset, there is a direct relationship between $n$ and the performance, up until 50 facial landmarks, from where the performance starts to drop. This behaviour can be justified as the $n$ increases, the overlap between the
patches around these landmarks, and thus the redundant information increases. Almost the same behaviour can be seen for the FER2013 dataset. To
be consistent in all the experiments, we set the value of $n$ as 50.

\subsection{Optimal Patch Size} \label{sec:PatchSize}
As mentioned earlier in~\ref{sec:TextureStream}, we apply $a \times a$ patches over each of the extracted facial landmarks, and subsequently feed these patches to the encoder to extract texture information. In this experiment, we aim to find the optimal value for the parameter $a$. To this end, we first consider 4 $\times$ 4 patches for training FaceTopoNet on both AffectNet and FER2013 datasets. Next, we repeat the experiment by increasing $a$ by 1. We continue this process until no performance increase is observed. Figure~\ref{fig:OptPatch} depicts the changes in RR with respect to the changes in $a$ for both datasets. This figure shows that the highest RR for AffectNet and FER2013 occurs at $a=17$ and $a=16$ respectively. 
%This figure demonstrates that the optimal patch size for AffectNet and FER2013 are respectively $a = 17$, and $a = 16$, resulting in RRs of 70.02\% and 72.9\%. 
The small difference between the optimal patch sizes of the two datasets indicates low dependency of $a$ on the dataset.
% -that the optimal value of $a$ does not depend on the datasets. 
For consistency, we set $a=17$ throughout our experimental setup.
% as the selected value for the parameter $a$ . For consistency, all the results reported in this paper are obtained by setting $a$ as 17.

\begin{figure}
    \begin{center}
    \includegraphics[width=0.7\columnwidth]{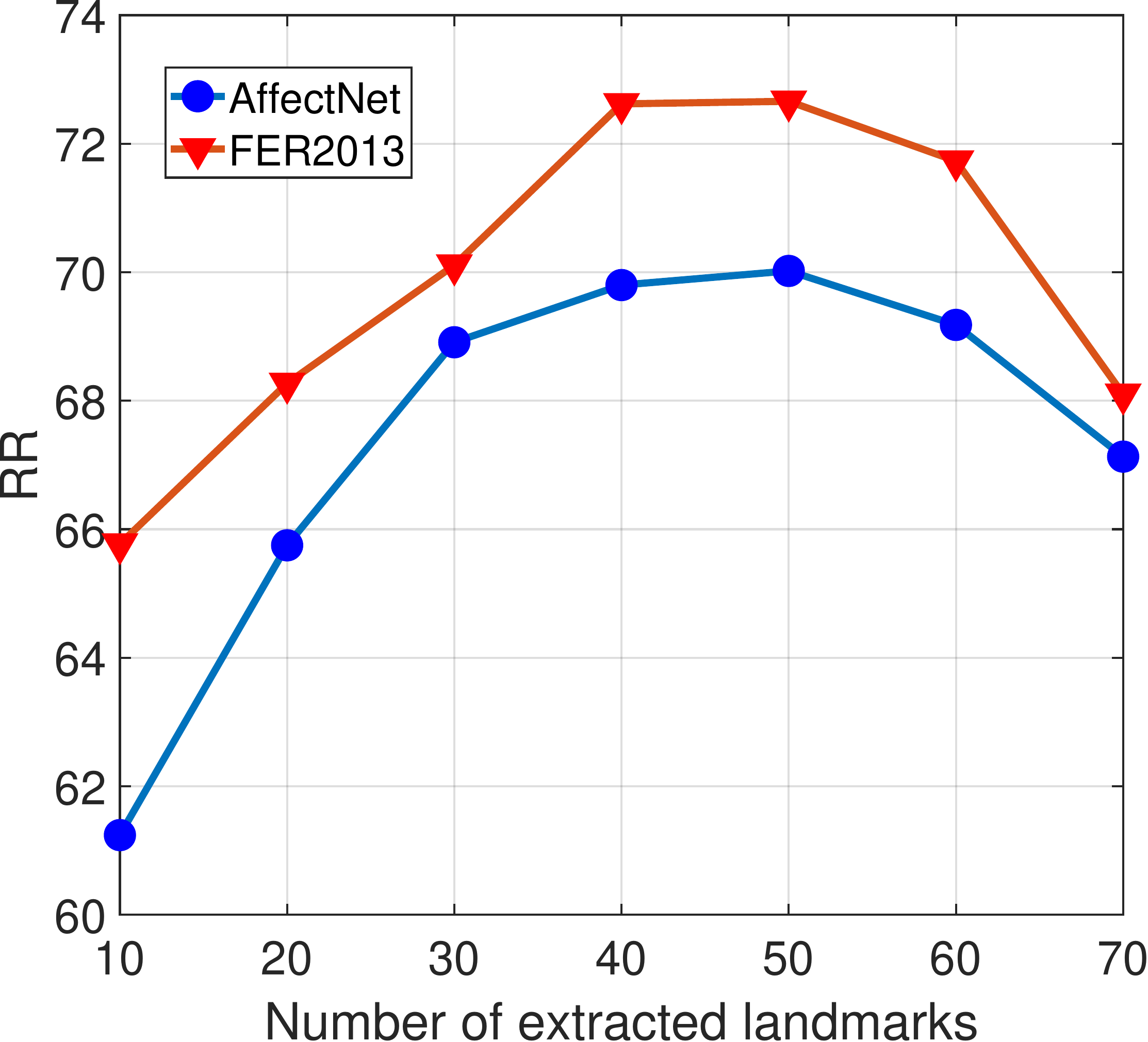}
    \end{center}
\caption{Changes in RR with the changes in the number of extracted landmarks}
\label{fig:OptFacialLandmraks}
\end{figure}

\begin{figure}
    \begin{center}
    \includegraphics[width=0.7\columnwidth]{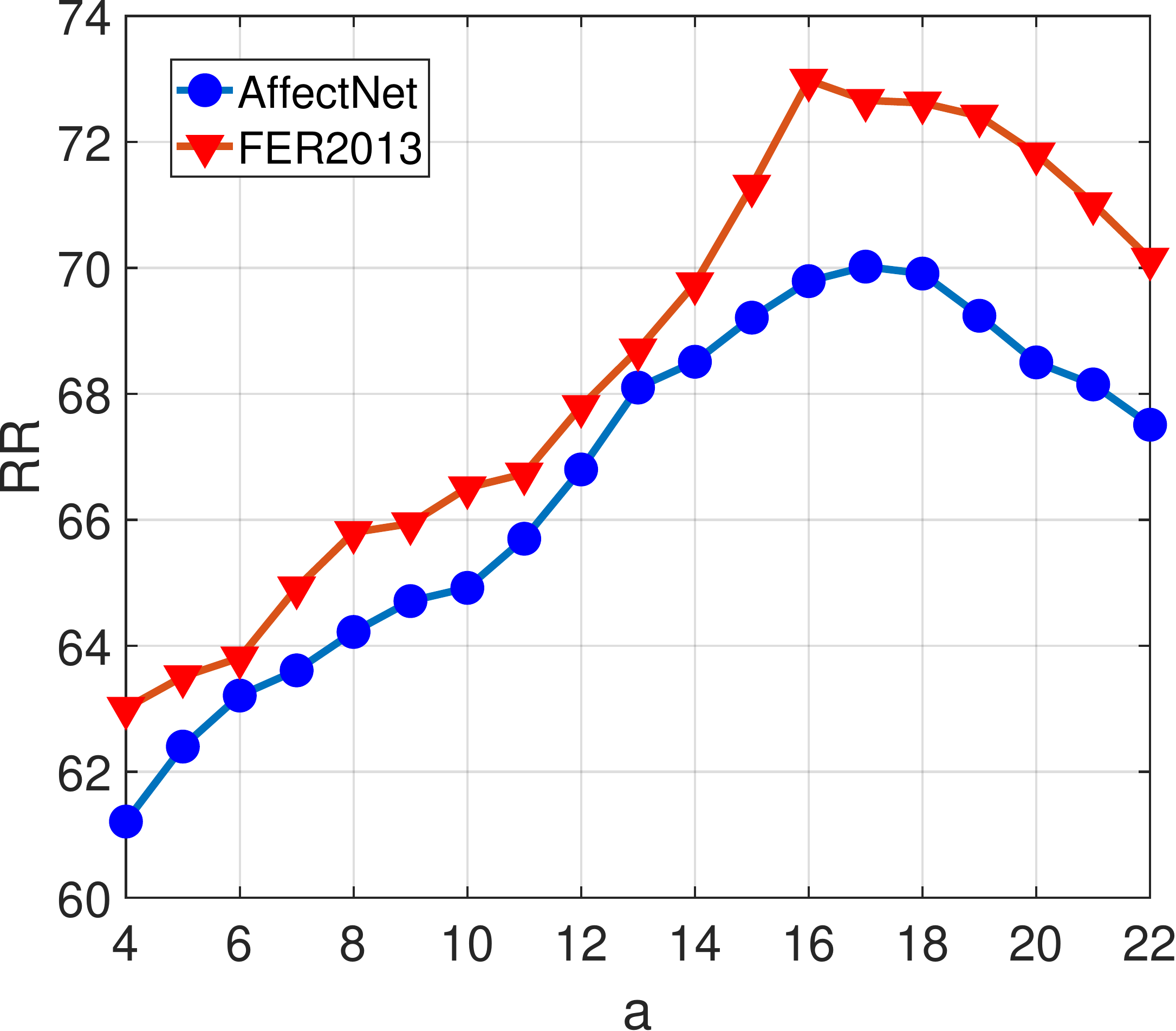}
    \end{center}
\caption{Changes in RR with the changes in $a$.}
\label{fig:OptPatch}
\end{figure}

\subsection{Robustness Occlusion}
\begin{figure*}
    \begin{center}
    \includegraphics[scale=0.10]{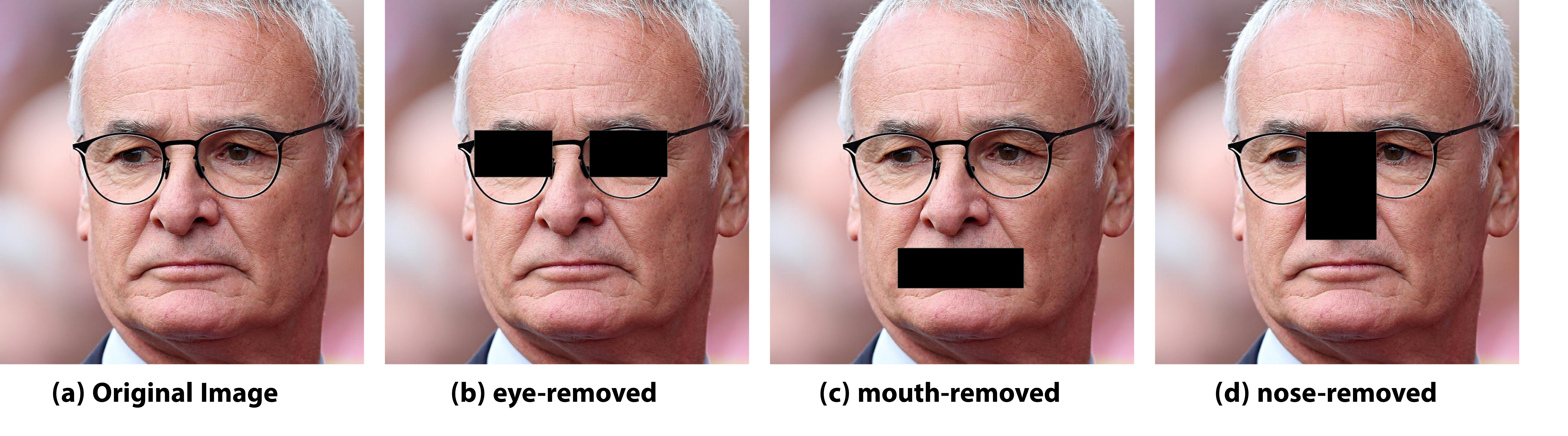}
    \end{center}
\caption{Added occlusions on an image from AffectNet dataset}
\label{fig:occlusion}
\end{figure*}

\begin{figure}
    \begin{center}
    \includegraphics[width=\columnwidth]{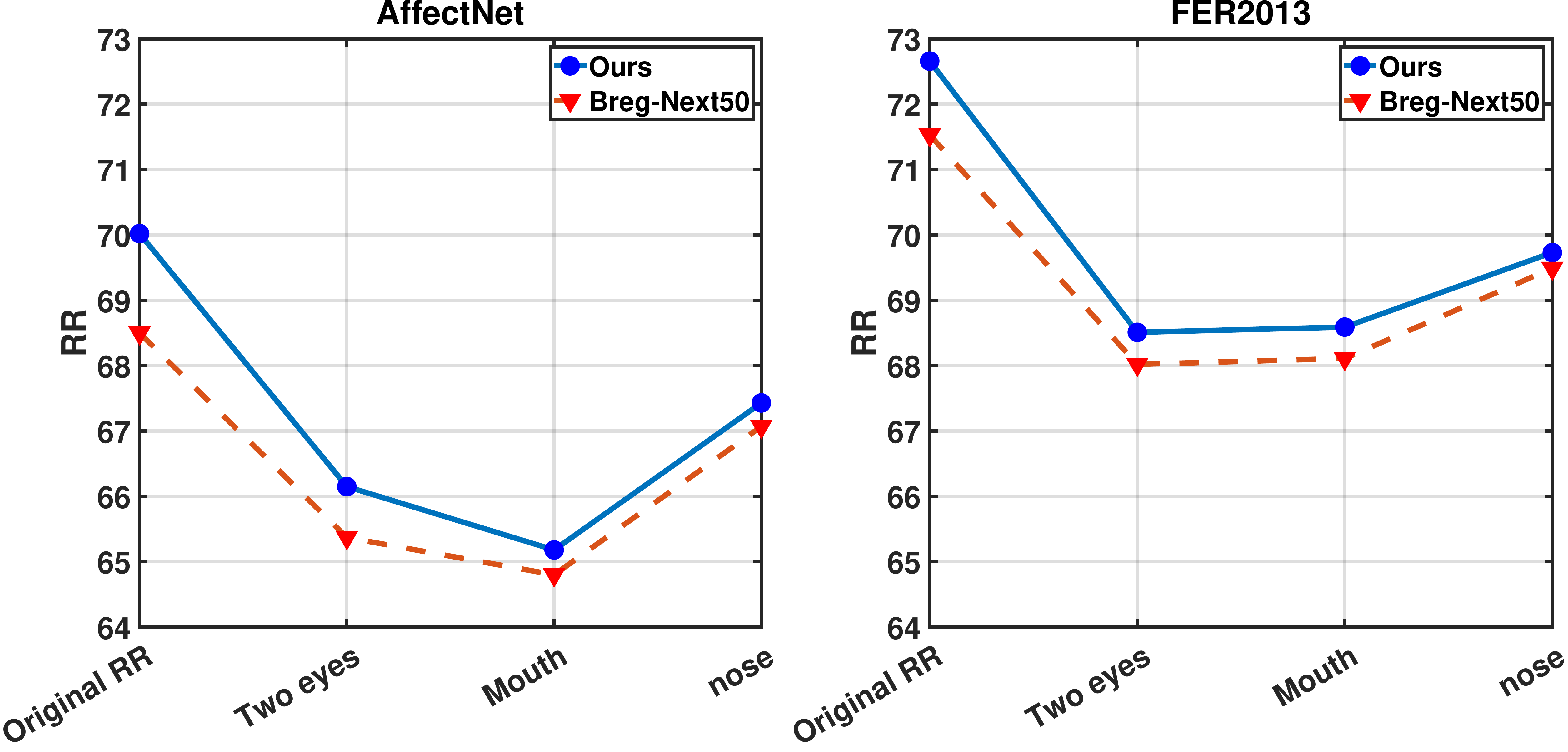}
    \end{center}
\caption{Changes of RR with regards to different types of occlusions}
\label{fig:occlusionResult}
\end{figure}

Here, we evaluate the robustness of FaceTopoNet and the state-of-the-art BreG-NeXt-50 against synthetic occlusions placed on key areas of the face. Considering the availability of the state-of-the-art pre-trained models on AffectNet and FER2013, we decide to perform this experiment on these two datasets. To this end, we extract the key components of face images, namely two eyes, mouth, and nose, using OpenCV \cite{bradski2000opencv} and dlib~\cite{dlib09} libraries. 
% Dlib is a popular library among researchers for extraction of facial components and it has been widely used. 
We then add black patches on these key areas, and subsequently build three new validation sets for each of the AffectNet and FER2013 datasets to test our model against. We name these validation sets ``eye-removed'', ``mouth-removed'', and ``nose-removed''. Figure~\ref{fig:occlusion} shows a sample image along with the associated occlusions. 
% We test FaceTopoNet and the state-of-the-art BreG-NeXt50 method with the new "eye-removed", "mouth-removed", and "nose-removed" validation sets. 
The performance of our model on both datasets is shown in Figure~\ref{fig:occlusionResult}, where FaceTopoNet delivers better RRs (0.6\% average gain) compared to  Breg-NeXt-50 in all the occlusion scenarios. This demonstrates the robustness of our method against the considered occlusions. Additionally, this experiment reveals that mouth and eyes have approximately the same contribution in expression recognition, since ``eye-removed'' and ``mouth-removed'' scenarios show roughly the same RR drop. This finding is in line with previous studies such as~\cite{koch2005role}. 

\begin{figure}[t]
    \begin{center}
    \includegraphics[width=\columnwidth]{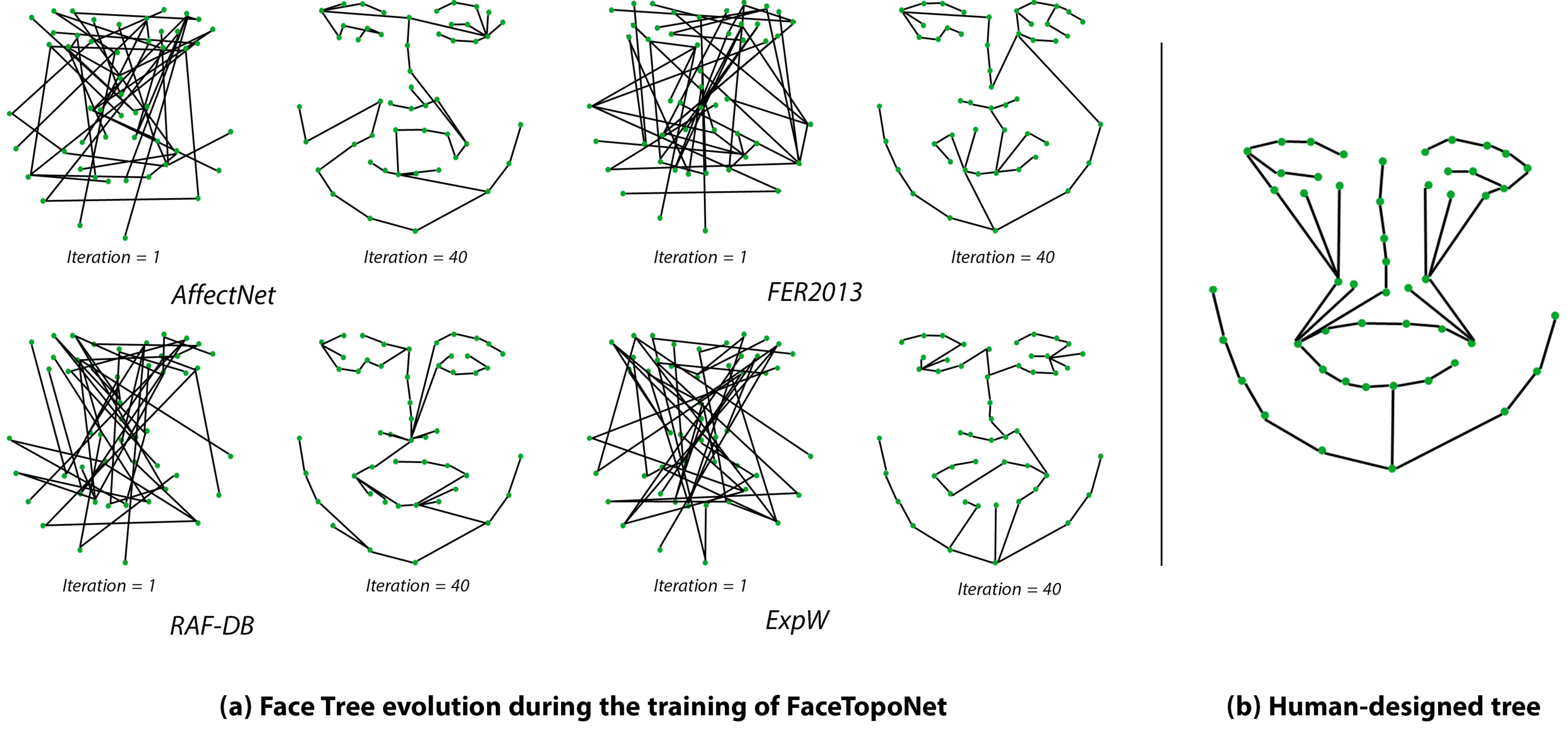}
    \end{center}
\caption{Face trees obtained by FaceTopoNet and Human-designed face tree borrowed from~\cite{G1}}
\label{fig:treeEvol}
\end{figure}

\subsection{Tree Topology Evolution}
The evolution of the topology of the face trees in the training phase of FaceTopoNet can provide some additional insights into the learning process. Figure~\ref{fig:treeEvol} depicts the topological evolution of the minimum-cost spanning trees generated by FaceTopoNet in iterations 1 and 40 for AffectNet, FER2013, RAF-DB, and ExpW datasets. In the beginning of the training phase, as the weights of $K_n$ are random, the trees are also random. However, as the training progresses, the face trees become more resembling of the structure of a human face. Moreover, from the face trees created in iteration 40, it is evident that there exists variations between the face tree created for each dataset, particularly around nose and mouth regions. This is not unexpected given the importance of these regions in FER.
%In the beginning of the training phase, FaceTopoNet creates a random tree. This is done by adding random weights to the edges of the complete graph. It is obvious that face trees become more resembling of the structure of a human face as the training process progresses. For example, face trees which are created in iteration 40 for all the datasets approximate perceivable human faces, although they are also customized for each dataset. This resemblance is more obvious around the nose, eyes, jaw line, and mouth. Moreover, from the face trees created in iteration 40, it is evident that there exists variations between the face tree created for each dataset, particularly around nose and mouth regions. This is not unexpected given the importance of these regions in FER. 
%As a result, our proposed FaceTopoNet tries to learn the more informative regions of the face based on the specifications of each dataset.

% \begin{figure*}[t]
%     \begin{center}
%     % \includegraphics[width=0.74\textwidth]{fig3j.pdf}
%     % \includegraphics[width=0.80\textwidth]{fig3-4dataset-v3.pdf}
%     \includegraphics[width=0.50\textwidth]{fig3-4dataset-v4.pdf}
%     \end{center}
% \caption{Face trees obtained by FaceTopoNet and Human-designed face tree borrowed from~\cite{G1}}
% \label{fig:treeEvol}
% \end{figure*}

\subsection{Cross-Dataset Study and Comparison to Human-Designed Face Tree}
% Transfer learning seeks to improve target learners' performance defined on target domains by knowledge transformation and it has been a promising area in machine learning~\cite{zhuang2020comprehensive}. 
Here, we aim to explore \textit{whether or not the trees generated by FaceTopoNet can be used in transfer learning}. To this end, we perform a cross-dataset study
% . to evaluate that FaceTopoNet is capable of doing transfer learning. 
% Moreover, in the second part, we want to show that trees formed by FaceTopoNet outperforms human-designed face trees.
% For conducting the first experiment, 
by using the final tree originally learned using AffectNet for expression recognition on the FER2013 dataset, and vice-versa. 
% Likewise, the tree obtained for AffectNet is adopted for FER2013 dataset (see the trees in the figure~\ref{fig:treeEvol}). 
% Utilizing these two trees which are used instead of one another, we conduct FER for the images inside the validation set of AffectNet and FER2013 datasets. 
Table~\ref{table:cross-dataset} presents the results where we observe an RR drop of 1.54\% and 1.26\% for AffectNet and FER2013 datasets, respectively. From this experiment, we conclude that while the learned trees are customized and optimized for individual datasets, given the similarities between the datasets, it is reasonably feasible to perform cross-dataset tree-based transfer learning.
% Despite this drop in RR, the results are still comparable to the state-of-the-art, which demonstrate the superiority of FaceTopoNet. For doing the transfer learning and seeing the results, we go further by fine tuning those two trees which are used instead of each other. It is observed that the trees are converged to their original topology. This matter prove that transfer learning works in the trees generated by FaceTopoNet.

Next we compare the trees generated by FaceTopoNet with human-designed trees which have been used in other literature~\cite{G1} (see Figure~\ref{fig:treeEvol} (b)). This tree is built based on the physiological as and psychological properties of human faces. Thus, we use this tree to perform FER on AffectNet and FER2013 datasets instead of using the customized trees learned for each individual dataset. Table~\ref{table:cross-dataset} illustrates the results where we observe a drop of 4.11\% and 5.4\% for AffectNet and FER2013 datasets, respectively. We can therefore infer that FaceTopoNet-generated trees are more optimized for conducting FER for various datasets in comparison to human-designed trees.% and there exists a considerable performance difference between the two.

\begin{table}
\centering
\caption{Cross-dataset study}
\label{table:cross-dataset}
\begin{tabular}{l|l|l} 
\hline
\textbf{Dataset} & \textbf{Tree}  & \textbf{RR}  \\ 
\hline
AffectNet        & FER2013        & 68.48        \\
FER2013          & AffectNet      & 71.40        \\
AffectNet        & Human-designed~\cite{G1} & 65.91        \\
FER2013          & Human-designed~\cite{G1} & 67.26        \\
\hline
\end{tabular}
\end{table}

% \begin{figure}[t]
%     \begin{center}
%     \includegraphics[scale=0.18]{human-designed-tree.pdf}
%     \end{center}
% \caption{Human-designed tree borrowed from~\cite{G1}.}
% \label{fig:human-designed-tree}
% \end{figure}

\subsection{Ablation study}
We conduct ablation experiments to evaluate the contribution of different components of our pipeline. This is done by systematic removal of the components of FaceTopoNet and calculating the resulting drop in performance. We present the results of this study in Table~\ref{table:abblation}, where (\textit{i}) the tree topology learning step is removed and replaced by a random tree since the model requires a tree; (\textit{ii}) the texture stream is removed; (\textit{iii}) the structure stream is removed; and (\textit{iv}) the final fusion step is removed and replaced by a simple concatenation. From the results outlined in Table~\ref{table:abblation}, we observe that the removal of the tree topology learning step results in substantial performance drop of 5.0\% and 7.60\%, respectively for AffectNet and FER2013 datasets. Moreover, it is shown that while both texture and structure streams are critical toward the final performance, the texture stream plays a more critical role. This can be attributed to the higher richness of information in texture patches versus structure coordinates.
% because of the use of pre-trained ResNet50 in the texture stream, the contribution of the texture stream to the FaceTopoNet is more significant than the structure stream. 
Lastly the model also exhibits a drop in performance when the two-stream fusion strategy is replaced with a simple feature-level 
% ??? 
fusion, revealing the importance of conducting a proper fusion in the final step of FaceTopoNet. 
% Additionally, relative contribution of the fusion is less than the tree topology learning, which indicates the importance of topology learning in the pipeline.

\begin{table}
\begin{center}
\caption{Ablation study. Recognition rates (RR) and the resulting drop in RR are provided when a specific component is removed from FaceTopoNet}
\label{table:abblation}
\footnotesize
\begin{tabular}{c|c|c|c|c}
\hline
\textbf{Removed} & \multicolumn{2}{|c|}{\textbf{RR}} &  \multicolumn{2}{|c}{\textbf{Drop in RR}} \\
\cline{2-5}
 & \textbf{AffectNet} & \textbf{FER2013} & \textbf{AffectNet} & \textbf{FER2013}\\
 \hline
 \hline
Tree topology & 66.51\% & 67.11\% & 5.0\% & 7.60\%\\
Structure stream & 64.90\% & 66.02\% & 7.30\% & 9.14\%\\
Texture stream & 62.14\% & 63.17\% & 11.25\% & 13.06\%\\
Fusion strategy & 66.87\% & 68.24\% & 4.50\% & 6.08\%\\
\hline
\end{tabular}
\end{center}
\end{table}

\subsection{Limitations}
%As FaceTopoNet includes two different streams, tree traversal, minimum-cost spanning tree calculation, and requires extracting facial landmarks, we perform an experiment to estimate its inference time and compare it with that of the others. To this end, we apply FaceTopoNet over 100 test images and average the corresponding inference time. Also, we find~\cite{wang2020region} and~\cite{kuo2018compact} which report their inference time in the context of FER. From Table~\ref{table:inference time}, it is evident that FaceTopoNet's run-time is higher, but by a small margin. One of the ways by which FaceTopoNet can be improved is using faster algorithms for finding the minimum-cost spanning tree.
%We perform an experiment to estimate the inference time of our method and compare it with that of prior work. To this end, we apply FaceTopoNet over 100 test images and average the corresponding inference time. Also, we find~\cite{wang2020region} and~\cite{kuo2018compact} which report their inference time in the context of FER. To be fair, we re-run their method on our local machine with the same hardware used to record the inference time of FaceTopoNet.  From Table~IX, it is evident that FaceTopoNet's run-time is higher, but by a small margin. One of the ways by which FaceTopoNet can be improved is using faster algorithms for the calculation of the minimum-cost spanning tree
We perform an experiment to estimate the inference time of our method and compare it with that of prior work. To this end, we apply FaceTopoNet over 100 test images and average the corresponding inference time. Also, we find~\cite{wang2020region} and~\cite{kuo2018compact} which report their inference time in the context of FER. For a fair comparison, we re-run their method on our local machine with the same hardware used to record the inference time of FaceTopoNet. From Table IX, it is evident that FaceTopoNet's run-time is higher, but only by a small margin. One of the ways by which FaceTopoNet can be improved is using faster algorithms for the calculation of the minimum-cost spanning tree.

\begin{table}
\centering

\caption{Inference time of different methods}
\label{table:inference time}
\begin{tabular}{l|l|l} 
\hline
\textbf{Author} & \textbf{Method}  & \textbf{Inference time (ms)}  \\ 
\hline
Wang et al.~\cite{wang2020region}        & Region attention        & 24        \\
Kuo et al.\cite{kuo2018compact}          & Compact CNN      & 21        \\
Ours        & FaceTopoNet & 33        \\
\hline
\end{tabular}
\end{table}

\section{Conclusion} \label{sec:conclusion}
In this paper, we present a new end-to-end deep pipeline, termed FaceTopoNet, for learning the topology of face graphs for facial expression recognition systems. FaceTopoNet first extracts facial landmarks which are then used to create a fully connected graph. 
The weights of the graph are then optimized to create a minimum-cost spanning tree, which is then traversed to create a sequence. This sequence is subsequently used to design the input embeddings of two sequential learners in two parallel streams within our model.
The first stream relies on the coordinates of the landmarks to learn the structure of the face, while the second stream learns texture information using patches around each extracted landmark. The outputs of each stream are then passed through an attention mechanism to focus on more salient features, prior to being fused.
We perform extensive experiments on 4 popular in-the-wild FER datasets, AffectNet, FER2013, RAF-DB, and ExpW. Results indicate that not only FaceTopoNet sets new state-of-the-art recognition rate in AffectNet, FER2013, and ExpW datasets, but also shows more robustness against different occlusions. We also perform detailed ablation and sensitivity studies to study our model in depth.

Future work may include using new optimization approaches such as reinforcement learning for further optimization of the tree structure. Moreover, our approach may be used for optimization of the structure of graph convolutional neural networks used for face representation learning.

% the applies weights to the edges of a fully-connected graph which they create.
% reflecting all potential connections between facial landmarks. 
% The weights are subsequently included into an optimization procedure. 

% together with its weights, are utilised to create a minimum-cost spanning tree, the traversal of which results in an optimum sequence. 
% We make use of this sequence to form an embedding that will be fed into a sequential learner. 
% FaceTopoNet comprises 2 different pipelines. 
% Regarding this, facial landmark's spatial information are used in the first stream. 
% However, the second stream utilizes the texture information of the face images, by applying $n$ by $n$ patches around facial landmarks. For conducting FER at the final step, we make use of the two-stream fusion strategy to fuse the learned representations of each stream. For demonstrating the efficacy of the FaceTopoNet, 

%meta heuristic algorithms.

\bibliographystyle{IEEEtran}  
\bibliography{egbib}

\begin{thebibliography}{10}
\providecommand{\url}[1]{#1}
\csname url@samestyle\endcsname
\providecommand{\newblock}{\relax}
\providecommand{\bibinfo}[2]{#2}
\providecommand{\BIBentrySTDinterwordspacing}{\spaceskip=0pt\relax}
\providecommand{\BIBentryALTinterwordstretchfactor}{4}
\providecommand{\BIBentryALTinterwordspacing}{\spaceskip=\fontdimen2\font plus
\BIBentryALTinterwordstretchfactor\fontdimen3\font minus
  \fontdimen4\font\relax}
\providecommand{\BIBforeignlanguage}[2]{{%
\expandafter\ifx\csname l@#1\endcsname\relax
\typeout{** WARNING: IEEEtran.bst: No hyphenation pattern has been}%
\typeout{** loaded for the language `#1'. Using the pattern for}%
\typeout{** the default language instead.}%
\else
\language=\csname l@#1\endcsname
\fi
#2}}
\providecommand{\BIBdecl}{\relax}
\BIBdecl

\bibitem{jaques2016understanding}
N.~Jaques, D.~McDuff, Y.~L. Kim, and R.~Picard, ``Understanding and predicting
  bonding in conversations using thin slices of facial expressions and body
  language,'' in \emph{International Conference on Intelligent Virtual
  Agents}.\hskip 1em plus 0.5em minus 0.4em\relax Springer, 2016, pp. 64--74.

\bibitem{picard2000affective}
R.~W. Picard, \emph{Affective computing}.\hskip 1em plus 0.5em minus
  0.4em\relax MIT press, 2000.

\bibitem{sepas2021capsfield}
A.~Sepas-Moghaddam, A.~Etemad, F.~Pereira, and P.~L. Correia, ``Capsfield:
  light field-based face and expression recognition in the wild using capsule
  routing,'' \emph{arXiv preprint arXiv:2101.03503}, 2021.

\bibitem{shuvendu2021}
R.~Shuvendu and A.~Etemad, ``Self-supervised contrastive learning of multi-view
  facial expressions,'' in \emph{Proceedings of the 2021 International
  Conference on Multimodal Interaction}, 2021.

\bibitem{sepas2020facial}
A.~Sepas-Moghaddam, A.~Etemad, F.~Pereira, and P.~L. Correia, ``Facial emotion
  recognition using light field images with deep attention-based bidirectional
  lstm,'' in \emph{ICASSP 2020-2020 IEEE International Conference on Acoustics,
  Speech and Signal Processing (ICASSP)}.\hskip 1em plus 0.5em minus
  0.4em\relax IEEE, 2020, pp. 3367--3371.

\bibitem{sepas2019deep}
A.~Sepas-Moghaddam, A.~Etemad, P.~L. Correia, and F.~Pereira, ``A deep
  framework for facial emotion recognition using light field images,'' in
  \emph{2019 8th International Conference on Affective Computing and
  Intelligent Interaction (ACII)}.\hskip 1em plus 0.5em minus 0.4em\relax IEEE,
  2019, pp. 1--7.

\bibitem{bargshady2020enhanced}
G.~Bargshady, X.~Zhou, R.~C. Deo, J.~Soar, F.~Whittaker, and H.~Wang,
  ``Enhanced deep learning algorithm development to detect pain intensity from
  facial expression images,'' \emph{Expert Systems with Applications}, vol.
  149, p. 113305, 2020.

\bibitem{pei2019attended}
W.~Pei, H.~Dibeklio{\u{g}}lu, T.~Baltru{\v{s}}aitis, and D.~M. Tax, ``Attended
  end-to-end architecture for age estimation from facial expression videos,''
  \emph{IEEE Transactions on Image Processing}, vol.~29, pp. 1972--1984, 2019.

\bibitem{porcu2020estimation}
S.~Porcu, A.~Floris, J.-N. Voigt-Antons, L.~Atzori, and S.~M{\"o}ller,
  ``Estimation of the quality of experience during video streaming from facial
  expression and gaze direction,'' \emph{IEEE Transactions on Network and
  Service Management}, vol.~17, no.~4, pp. 2702--2716, 2020.

\bibitem{li2020depression}
X.~Li, W.~Guo, and H.~Yang, ``Depression severity prediction from facial
  expression based on the drr\_depressionnet network,'' in \emph{2020 IEEE
  International Conference on Bioinformatics and Biomedicine (BIBM)}.\hskip 1em
  plus 0.5em minus 0.4em\relax IEEE, 2020, pp. 2757--2764.

\bibitem{surveyAsli}
S.~Li and W.~Deng, ``Deep facial expression recognition: A survey,'' \emph{IEEE
  Transactions on Affective Computing}, vol.~13, no.~3, pp. 1195--1215, 2022.

\bibitem{jung2015joint}
H.~Jung, S.~Lee, J.~Yim, S.~Park, and J.~Kim, ``Joint fine-tuning in deep
  neural networks for facial expression recognition,'' in \emph{Proceedings of
  the IEEE international conference on computer vision}, 2015, pp. 2983--2991.

\bibitem{ghimire2017facial}
D.~Ghimire, S.~Jeong, J.~Lee, and S.~H. Park, ``Facial expression recognition
  based on local region specific features and support vector machines,''
  \emph{Multimedia Tools and Applications}, vol.~76, no.~6, pp. 7803--7821,
  2017.

\bibitem{happy2020expression}
S.~Happy, A.~Dantcheva, and F.~Bremond, ``Expression recognition with deep
  features extracted from holistic and part-based models,'' \emph{Image and
  Vision Computing}, p. 104038, 2020.

\bibitem{liu2019hard}
X.~Liu, B.~V. Kumar, P.~Jia, and J.~You, ``Hard negative generation for
  identity-disentangled facial expression recognition,'' \emph{Pattern
  Recognition}, vol.~88, pp. 1--12, 2019.

\bibitem{yu2020facial}
M.~Yu, H.~Zheng, Z.~Peng, J.~Dong, and H.~Du, ``Facial expression recognition
  based on a multi-task global-local network,'' \emph{Pattern Recognition
  Letters}, vol. 131, pp. 166--171, 2020.

\bibitem{GCN1}
H.-X. Xie, L.~Lo, H.-H. Shuai, and W.-H. Cheng, ``Au-assisted graph attention
  convolutional network for micro-expression recognition,'' in \emph{ACM
  International Conference on Multimedia}, 2020, pp. 2871--2880.

\bibitem{GCN2}
Y.~Xie, T.~Chen, T.~Pu, H.~Wu, and L.~Lin, ``Adversarial graph representation
  adaptation for cross-domain facial expression recognition,'' in \emph{ACM
  International Conference on Multimedia}, 2020, pp. 1255--1264.

\bibitem{GRNN}
L.~Zhong, C.~Bai, J.~Li, T.~Chen, S.~Li, and Y.~Liu, ``A graph-structured
  representation with brnn for static-based facial expression recognition,'' in
  \emph{IEEE International Conference on Automatic Face \& Gesture
  Recognition}, 2019, pp. 1--5.

\bibitem{G1}
J.~Zhou, X.~Zhang, Y.~Liu, and X.~Lan, ``Facial expression recognition using
  spatial-temporal semantic graph network,'' in \emph{2020 IEEE International
  Conference on Image Processing (ICIP)}.\hskip 1em plus 0.5em minus
  0.4em\relax IEEE, 2020, pp. 1961--1965.

\bibitem{G2}
L.~Lei, J.~Li, T.~Chen, and S.~Li, ``A novel graph-tcn with a graph structured
  representation for micro-expression recognition,'' in \emph{ACM International
  Conference on Multimedia}, 2020, pp. 2237--2245.

\bibitem{G3}
X.~Xu, Z.~Ruan, and L.~Yang, ``Facial expression recognition based on graph
  neural network,'' in \emph{International Conference on Image, Vision and
  Computing}, 2020, pp. 211--214.

\bibitem{liu2019facial}
Y.~Liu, X.~Zhang, Y.~Lin, and H.~Wang, ``Facial expression recognition via deep
  action units graph network based on psychological mechanism,'' \emph{IEEE
  Transactions on Cognitive and Developmental Systems}, vol.~12, no.~2, pp.
  311--322, 2019.

\bibitem{li2011cooperatively}
X.~Li and X.~Yao, ``Cooperatively coevolving particle swarms for large scale
  optimization,'' \emph{IEEE Transactions on Evolutionary Computation},
  vol.~16, no.~2, pp. 210--224, 2011.

\bibitem{Affectnet}
A.~Mollahosseini, B.~Hasani, and M.~H. Mahoor, ``Affectnet: A database for
  facial expression, valence, and arousal computing in the wild,'' \emph{IEEE
  Transactions on Affective Computing}, vol.~10, no.~1, pp. 18--31, 2017.

\bibitem{FER2013}
I.~J. Goodfellow, D.~Erhan, P.~L. Carrier, A.~Courville, M.~Mirza, B.~Hamner,
  W.~Cukierski, Y.~Tang, D.~Thaler, D.-H. Lee \emph{et~al.}, ``Challenges in
  representation learning: A report on three machine learning contests,'' in
  \emph{International Conference on Neural Information Processing}, 2013, pp.
  117--124.

\bibitem{li2017reliable}
S.~Li, W.~Deng, and J.~Du, ``Reliable crowdsourcing and deep
  locality-preserving learning for expression recognition in the wild,'' in
  \emph{2017 IEEE Conference on Computer Vision and Pattern Recognition
  (CVPR)}.\hskip 1em plus 0.5em minus 0.4em\relax IEEE, 2017, pp. 2584--2593.

\bibitem{SOCIALRELATION_2017}
Z.~Zhang, P.~Luo, C.~C. Loy, and X.~Tang, ``From facial expression recognition
  to interpersonal relation prediction,'' in \emph{arXiv:1609.06426v2},
  September 2016.

\bibitem{kolahdouzi2021face}
M.~Kolahdouzi, A.~Sepas-Moghaddam, and A.~Etemad, ``Face trees for expression
  recognition,'' in \emph{2021 16th IEEE International Conference on Automatic
  Face and Gesture Recognition (FG 2021)}, 2021, pp. 1--5.

\bibitem{tang2013deep}
Y.~Tang, ``Deep learning using linear support vector machines,'' \emph{arXiv
  preprint arXiv:1306.0239}, 2013.

\bibitem{yao2016holonet}
A.~Yao, D.~Cai, P.~Hu, S.~Wang, L.~Sha, and Y.~Chen, ``Holonet: towards robust
  emotion recognition in the wild,'' in \emph{Proceedings of the 18th ACM
  International Conference on Multimodal Interaction}, 2016, pp. 472--478.

\bibitem{shang2016understanding}
W.~Shang, K.~Sohn, D.~Almeida, and H.~Lee, ``Understanding and improving
  convolutional neural networks via concatenated rectified linear units,'' in
  \emph{International Conference on Machine Learning}, 2016, pp. 2217--2225.

\bibitem{sang2017facial}
D.~V. Sang, N.~Van~Dat \emph{et~al.}, ``Facial expression recognition using
  deep convolutional neural networks,'' in \emph{International Conference on
  Knowledge and Systems Engineering}, 2017, pp. 130--135.

\bibitem{chen2019facial}
Y.~Chen, J.~Wang, S.~Chen, Z.~Shi, and J.~Cai, ``Facial motion prior networks
  for facial expression recognition,'' in \emph{IEEE Visual Communications and
  Image Processing}, 2019.

\bibitem{wang2020region}
K.~Wang, X.~Peng, J.~Yang, D.~Meng, and Y.~Qiao, ``Region attention networks
  for pose and occlusion robust facial expression recognition,'' \emph{IEEE
  Transactions on Image Processing}, vol.~29, pp. 4057--4069, 2020.

\bibitem{jiang2020accurate}
P.~Jiang, G.~Liu, Q.~Wang, and J.~Wu, ``Accurate and reliable facial expression
  recognition using advanced softmax loss with fixed weights,'' \emph{IEEE
  Signal Processing Letters}, vol.~27, pp. 725--729, 2020.

\bibitem{kollias2020deep}
D.~Kollias, S.~Cheng, E.~Ververas, I.~Kotsia, and S.~Zafeiriou, ``Deep neural
  network augmentation: Generating faces for affect analysis,''
  \emph{International Journal of Computer Vision}, pp. 1--30, 2020.

\bibitem{booth20173d}
J.~Booth, E.~Antonakos, S.~Ploumpis, G.~Trigeorgis, Y.~Panagakis, and
  S.~Zafeiriou, ``3d face morphable models" in-the-wild",'' in
  \emph{Proceedings of the IEEE Conference on Computer Vision and Pattern
  Recognition}, 2017, pp. 48--57.

\bibitem{zhao2021learning}
Z.~Zhao, Q.~Liu, and S.~Wang, ``Learning deep global multi-scale and local
  attention features for facial expression recognition in the wild,''
  \emph{IEEE Transactions on Image Processing}, vol.~30, pp. 6544--6556, 2021.

\bibitem{xue2021transfer}
F.~Xue, Q.~Wang, and G.~Guo, ``Transfer: Learning relation-aware facial
  expression representations with transformers,'' in \emph{Proceedings of the
  IEEE/CVF International Conference on Computer Vision}, 2021, pp. 3601--3610.

\bibitem{cai2021identity}
J.~Cai, Z.~Meng, A.~S. Khan, J.~O’Reilly, Z.~Li, S.~Han, and Y.~Tong,
  ``Identity-free facial expression recognition using conditional generative
  adversarial network,'' in \emph{2021 IEEE International Conference on Image
  Processing (ICIP)}.\hskip 1em plus 0.5em minus 0.4em\relax IEEE, 2021, pp.
  1344--1348.

\bibitem{ruan2021feature}
D.~Ruan, Y.~Yan, S.~Lai, Z.~Chai, C.~Shen, and H.~Wang, ``Feature decomposition
  and reconstruction learning for effective facial expression recognition,'' in
  \emph{Proceedings of the IEEE/CVF Conference on Computer Vision and Pattern
  Recognition}, 2021, pp. 7660--7669.

\bibitem{huang2021identity}
W.~Huang, S.~Zhang, P.~Zhang, Y.~Zha, Y.~Fang, and Y.~Zhang, ``Identity-aware
  facial expression recognition via deep metric learning based on synthesized
  images,'' \emph{IEEE Transactions on Multimedia}, 2021.

\bibitem{choi2018stargan}
Y.~Choi, M.~Choi, M.~Kim, J.-W. Ha, S.~Kim, and J.~Choo, ``Stargan: Unified
  generative adversarial networks for multi-domain image-to-image
  translation,'' in \emph{Proceedings of the IEEE conference on computer vision
  and pattern recognition}, 2018, pp. 8789--8797.

\bibitem{bisogni2022impact}
C.~Bisogni, A.~Castiglione, S.~Hossain, F.~Narducci, and S.~Umer, ``Impact of
  deep learning approaches on facial expression recognition in healthcare
  industries,'' \emph{IEEE Transactions on Industrial Informatics}, vol.~18,
  no.~8, pp. 5619--5627, 2022.

\bibitem{hossain2021unified}
S.~Hossain, S.~Umer, V.~Asari, and R.~K. Rout, ``A unified framework of deep
  learning-based facial expression recognition system for diversified
  applications,'' \emph{Applied Sciences}, vol.~11, no.~19, p. 9174, 2021.

\bibitem{umer2022facial}
S.~Umer, R.~K. Rout, C.~Pero, and M.~Nappi, ``Facial expression recognition
  with trade-offs between data augmentation and deep learning features,''
  \emph{Journal of Ambient Intelligence and Humanized Computing}, vol.~13,
  no.~2, pp. 721--735, 2022.

\bibitem{oztel2018ifer}
I.~Oztel, G.~Yolcu, C.~{\"O}z, S.~Kazan, and F.~Bunyak, ``ifer: facial
  expression recognition using automatically selected geometric eye and eyebrow
  features,'' \emph{Journal of Electronic Imaging}, vol.~27, no.~2, p. 023003,
  2018.

\bibitem{hasani2019bounded}
B.~Hasani, P.~S. Negi, and M.~H. Mahoor, ``Bounded residual gradient networks
  (breg-net) for facial affect computing,'' in \emph{2019 14th IEEE
  International Conference on Automatic Face \& Gesture Recognition (FG
  2019)}.\hskip 1em plus 0.5em minus 0.4em\relax IEEE, 2019, pp. 1--7.

\bibitem{hasani2020breg}
B.~Hasani, P.~S. Negi, and M.~Mahoor, ``Breg-next: facial affect computing
  using adaptive residual networks with bounded gradient,'' \emph{IEEE
  Transactions on Affective Computing}, vol.~13, no.~2, p.~1, 2022.

\bibitem{VQA}
D.~Teney, L.~Liu, and A.~van Den~Hengel, ``Graph-structured representations for
  visual question answering,'' in \emph{Proceedings of the IEEE Conference on
  Computer Vision and Pattern Recognition}, 2017, pp. 1--9.

\bibitem{ICLR}
Y.~Li, D.~Tarlow, M.~Brockschmidt, and R.~Zemel, ``Gated graph sequence neural
  networks,'' \emph{arXiv preprint arXiv:1511.05493}, 2015.

\bibitem{antoniadis2021exploiting}
P.~Antoniadis, P.~P. Filntisis, and P.~Maragos, ``Exploiting emotional
  dependencies with graph convolutional networks for facial expression
  recognition,'' \emph{arXiv preprint arXiv:2106.03487}, 2021.

\bibitem{liu2021video}
D.~Liu, H.~Zhang, and P.~Zhou, ``Video-based facial expression recognition
  using graph convolutional networks,'' in \emph{2020 25th International
  Conference on Pattern Recognition (ICPR)}.\hskip 1em plus 0.5em minus
  0.4em\relax IEEE, 2021, pp. 607--614.

\bibitem{Zhou2020facial}
J.~Zhou, X.~Zhang, Y.~Liu, and X.~Lan, ``Facial expression recognition using
  spatial-temporal semantic graph network,'' in \emph{2020 IEEE International
  Conference on Image Processing (ICIP)}.\hskip 1em plus 0.5em minus
  0.4em\relax IEEE, 2020, pp. 1961--1965.

\bibitem{liu2014deeply}
M.~Liu, S.~Li, S.~Shan, R.~Wang, and X.~Chen, ``Deeply learning deformable
  facial action parts model for dynamic expression analysis,'' in \emph{Asian
  Conference on Computer Vision}.\hskip 1em plus 0.5em minus 0.4em\relax
  Springer, 2014, pp. 143--157.

\bibitem{liu2020video}
D.~Liu, H.~Zhang, and P.~Zhou, ``Video-based facial expression recognition
  using graph convolutional networks,'' \emph{arXiv preprint arXiv:2010.13386},
  2020.

\bibitem{cho2014learning}
K.~Cho, B.~Van~Merri{\"e}nboer, C.~Gulcehre, D.~Bahdanau, F.~Bougares,
  H.~Schwenk, and Y.~Bengio, ``Learning phrase representations using rnn
  encoder-decoder for statistical machine translation,'' \emph{arXiv preprint
  arXiv:1406.1078}, 2014.

\bibitem{lv2017deep}
J.~Lv, X.~Shao, J.~Xing, C.~Cheng, and X.~Zhou, ``A deep regression
  architecture with two-stage re-initialization for high performance facial
  landmark detection,'' in \emph{Proceedings of the IEEE Conference on Computer
  Vision and Pattern Recognition}, 2017, pp. 3317--3326.

\bibitem{jiang2009research}
B.~Jiang and L.~Zhang, ``Research on minimum spanning tree based on prim
  algorithm,'' \emph{Computer Engineering and Design}, vol.~13, 2009.

\bibitem{kozen1992depth}
D.~C. Kozen, ``Depth-first and breadth-first search,'' in \emph{The Design and
  Analysis of Algorithms}.\hskip 1em plus 0.5em minus 0.4em\relax Springer,
  1992, pp. 19--24.

\bibitem{gers2000recurrent}
F.~A. Gers and J.~Schmidhuber, ``Recurrent nets that time and count,'' in
  \emph{Proceedings of the IEEE-INNS-ENNS International Joint Conference on
  Neural Networks. IJCNN 2000. Neural Computing: New Challenges and
  Perspectives for the New Millennium}, vol.~3.\hskip 1em plus 0.5em minus
  0.4em\relax IEEE, 2000, pp. 189--194.

\bibitem{rocktaschel2015reasoning}
T.~Rockt{\"a}schel, E.~Grefenstette, K.~M. Hermann, T.~Ko{\v{c}}isk{\`y}, and
  P.~Blunsom, ``Reasoning about entailment with neural attention,'' \emph{arXiv
  preprint arXiv:1509.06664}, 2015.

\bibitem{cao2018vggface2}
Q.~Cao, L.~Shen, W.~Xie, O.~M. Parkhi, and A.~Zisserman, ``Vggface2: A dataset
  for recognising faces across pose and age,'' in \emph{2018 13th IEEE
  International Conference on Automatic Face \& Gesture Recognition (FG
  2018)}.\hskip 1em plus 0.5em minus 0.4em\relax IEEE, 2018, pp. 67--74.

\bibitem{gu2018hybrid}
Y.~Gu, K.~Yang, S.~Fu, S.~Chen, X.~Li, and I.~Marsic, ``Hybrid attention based
  multimodal network for spoken language classification,'' in \emph{Proceedings
  of the Conference. Association for Computational Linguistics. Meeting}, vol.
  2018.\hskip 1em plus 0.5em minus 0.4em\relax NIH Public Access, 2018, p.
  2379.

\bibitem{zhang2020rfnet}
G.~Zhang and A.~Etemad, ``Rfnet: riemannian fusion network for eeg-based
  brain-computer interfaces,'' \emph{arXiv preprint arXiv:2008.08633}, 2020.

\bibitem{kingma2014adam}
D.~P. Kingma and J.~Ba, ``Adam: a method for stochastic optimization. arxiv,''
  \emph{arXiv preprint arXiv:1412.6980}, vol.~22, 2014.

\bibitem{he2009learning}
H.~He and E.~A. Garcia, ``Learning from imbalanced data,'' \emph{IEEE
  Transactions on Knowledge and Data Engineering}, vol.~21, no.~9, pp.
  1263--1284, 2009.

\bibitem{lin2017focal}
T.-Y. Lin, P.~Goyal, R.~Girshick, K.~He, and P.~Doll{\'a}r, ``Focal loss for
  dense object detection,'' in \emph{Proceedings of the IEEE International
  Conference on Computer Vision}, 2017, pp. 2980--2988.

\bibitem{abadi2016tensorflow}
M.~Abadi, P.~Barham, J.~Chen, Z.~Chen, A.~Davis, J.~Dean, M.~Devin,
  S.~Ghemawat, G.~Irving, M.~Isard \emph{et~al.}, ``Tensorflow: A system for
  large-scale machine learning,'' in \emph{12th $\{$USENIX$\}$ Symposium on
  Operating Systems Design and Implementation ($\{$OSDI$\}$ 16)}, 2016, pp.
  265--283.

\bibitem{vielzeuf2017temporal}
V.~Vielzeuf, S.~Pateux, and F.~Jurie, ``Temporal multimodal fusion for video
  emotion classification in the wild,'' in \emph{Proceedings of the 19th ACM
  International Conference on Multimodal Interaction}, 2017, pp. 569--576.

\bibitem{lucey2010extended}
P.~Lucey, J.~F. Cohn, T.~Kanade, J.~Saragih, Z.~Ambadar, and I.~Matthews, ``The
  extended cohn-kanade dataset (ck+): A complete dataset for action unit and
  emotion-specified expression,'' in \emph{2010 ieee computer society
  conference on computer vision and pattern recognition-workshops}.\hskip 1em
  plus 0.5em minus 0.4em\relax IEEE, 2010, pp. 94--101.

\bibitem{hewitt2018cnn}
C.~Hewitt and H.~Gunes, ``Cnn-based facial affect analysis on mobile devices,''
  \emph{arXiv preprint arXiv:1807.08775}, 2018.

\bibitem{hua2019hero}
W.~Hua, F.~Dai, L.~Huang, J.~Xiong, and G.~Gui, ``Hero: Human emotions
  recognition for realizing intelligent internet of things,'' \emph{IEEE
  Access}, vol.~7, pp. 24\,321--24\,332, 2019.

\bibitem{chen2021residual}
B.~Chen, W.~Guan, P.~Li, N.~Ikeda, K.~Hirasawa, and H.~Lu, ``Residual
  multi-task learning for facial landmark localization and expression
  recognition,'' \emph{Pattern Recognition}, vol. 115, p. 107893, 2021.

\bibitem{zhang2021learning}
W.~Zhang, X.~Ji, K.~Chen, Y.~Ding, and C.~Fan, ``Learning a facial expression
  embedding disentangled from identity,'' in \emph{Proceedings of the IEEE/CVF
  Conference on Computer Vision and Pattern Recognition}, 2021, pp. 6759--6768.

\bibitem{farzaneh2021facial}
A.~H. Farzaneh and X.~Qi, ``Facial expression recognition in the wild via deep
  attentive center loss,'' in \emph{Proceedings of the IEEE/CVF Winter
  Conference on Applications of Computer Vision}, 2021, pp. 2402--2411.

\bibitem{shi2021facial}
C.~Shi, C.~Tan, and L.~Wang, ``A facial expression recognition method based on
  a multibranch cross-connection convolutional neural network,'' \emph{IEEE
  Access}, vol.~9, pp. 39\,255--39\,274, 2021.

\bibitem{fard2022ad}
A.~P. Fard and M.~H. Mahoor, ``Ad-corre: Adaptive correlation-based loss for
  facial expression recognition in the wild,'' \emph{IEEE Access}, vol.~10, pp.
  26\,756--26\,768, 2022.

\bibitem{graves2013speech}
A.~Graves, A.-r. Mohamed, and G.~Hinton, ``Speech recognition with deep
  recurrent neural networks,'' in \emph{2013 IEEE international conference on
  acoustics, speech and signal processing}.\hskip 1em plus 0.5em minus
  0.4em\relax Ieee, 2013, pp. 6645--6649.

\bibitem{deng2009imagenet}
J.~Deng, W.~Dong, R.~Socher, L.-J. Li, K.~Li, and L.~Fei-Fei, ``Imagenet: A
  large-scale hierarchical image database,'' in \emph{2009 IEEE conference on
  computer vision and pattern recognition}.\hskip 1em plus 0.5em minus
  0.4em\relax Ieee, 2009, pp. 248--255.

\bibitem{rajan2020novel}
S.~Rajan, P.~Chenniappan, S.~Devaraj, and N.~Madian, ``Novel deep learning
  model for facial expression recognition based on maximum boosted cnn and
  lstm,'' \emph{IET Image Processing}, vol.~14, no.~7, pp. 1373--1381, 2020.

\bibitem{lian2020expression}
Z.~Lian, Y.~Li, J.-H. Tao, J.~Huang, and M.-Y. Niu, ``Expression analysis based
  on face regions in real-world conditions,'' \emph{International Journal of
  Automation and Computing}, vol.~17, no.~1, pp. 96--107, 2020.

\bibitem{li2021adaptively}
H.~Li, N.~Wang, X.~Ding, X.~Yang, and X.~Gao, ``Adaptively learning facial
  expression representation via cf labels and distillation,'' \emph{IEEE
  Transactions on Image Processing}, vol.~30, pp. 2016--2028, 2021.

\bibitem{galar2011overview}
M.~Galar, A.~Fern{\'a}ndez, E.~Barrenechea, H.~Bustince, and F.~Herrera, ``An
  overview of ensemble methods for binary classifiers in multi-class problems:
  Experimental study on one-vs-one and one-vs-all schemes,'' \emph{Pattern
  Recognition}, vol.~44, no.~8, pp. 1761--1776, 2011.

\bibitem{wang2021oaenet}
Z.~Wang, F.~Zeng, S.~Liu, and B.~Zeng, ``Oaenet: Oriented attention ensemble
  for accurate facial expression recognition,'' \emph{Pattern Recognition},
  vol. 112, p. 107694, 2021.

\bibitem{wang2020suppressing}
K.~Wang, X.~Peng, J.~Yang, S.~Lu, and Y.~Qiao, ``Suppressing uncertainties for
  large-scale facial expression recognition,'' in \emph{Proceedings of the
  IEEE/CVF Conference on Computer Vision and Pattern Recognition}, 2020, pp.
  6897--6906.

\bibitem{benamara2021real}
N.~K. Benamara, M.~Val-Calvo, J.~R. Alvarez-Sanchez, A.~D{\'\i}az-Morcillo,
  J.~M. Ferrandez-Vicente, E.~Fern{\'a}ndez-Jover, and T.~B. Stambouli,
  ``Real-time facial expression recognition using smoothed deep neural network
  ensemble,'' \emph{Integrated Computer-Aided Engineering}, no. Preprint, pp.
  1--15, 2021.

\bibitem{peng2022guided}
X.~Peng, Y.~Gu, and P.~Zhang, ``Au-guided unsupervised domain-adaptive facial
  expression recognition,'' \emph{Applied Sciences}, vol.~12, no.~9, p. 4366,
  2022.

\bibitem{ruan2020deep}
D.~Ruan, Y.~Yan, S.~Chen, J.-H. Xue, and H.~Wang, ``Deep
  disturbance-disentangled learning for facial expression recognition,'' in
  \emph{Proceedings of the 28th ACM International Conference on Multimedia},
  2020, pp. 2833--2841.

\bibitem{yang2018facial}
H.~Yang, U.~Ciftci, and L.~Yin, ``Facial expression recognition by
  de-expression residue learning,'' in \emph{Proceedings of the IEEE conference
  on computer vision and pattern recognition}, 2018, pp. 2168--2177.

\bibitem{zeng2018facialincon}
J.~Zeng, S.~Shan, and X.~Chen, ``Facial expression recognition with
  inconsistently annotated datasets,'' in \emph{Proceedings of the European
  Conference on Computer Vision (ECCV)}, 2018, pp. 222--237.

\bibitem{liu2017adaptive}
X.~Liu, B.~Vijaya~Kumar, J.~You, and P.~Jia, ``Adaptive deep metric learning
  for identity-aware facial expression recognition,'' in \emph{Proceedings of
  the IEEE conference on computer vision and pattern recognition workshops},
  2017, pp. 20--29.

\bibitem{mavani2017facial}
V.~Mavani, S.~Raman, and K.~P. Miyapuram, ``Facial expression recognition using
  visual saliency and deep learning,'' in \emph{Proceedings of the IEEE
  international conference on computer vision workshops}, 2017, pp. 2783--2788.

\bibitem{bradski2000opencv}
G.~Bradski and A.~Kaehler, ``Opencv,'' \emph{Dr. Dobb’s journal of software
  tools}, vol.~3, 2000.

\bibitem{dlib09}
D.~E. King, ``Dlib-ml: A machine learning toolkit,'' \emph{Journal of Machine
  Learning Research}, vol.~10, pp. 1755--1758, 2009.

\bibitem{koch2005role}
C.~Koch, ``The role of the eyes and mouth in facial emotions,'' in
  \emph{Abstracts of the Psychonomic Society}, vol.~10.\hskip 1em plus 0.5em
  minus 0.4em\relax Citeseer, 2005, p. 128.

\bibitem{kuo2018compact}
C.-M. Kuo, S.-H. Lai, and M.~Sarkis, ``A compact deep learning model for robust
  facial expression recognition,'' in \emph{Proceedings of the IEEE conference
  on computer vision and pattern recognition workshops}, 2018, pp. 2121--2129.

\end{thebibliography}

% \begin{IEEEbiography}{Third C. Author, Jr.}{\space}(M’87) received the B.S. degree in mechanical engineering from National Chung Cheng University, Chiayi, Taiwan, in 2004 and the M.S. degree in mechanical engineering from National Tsing Hua University, Hsinchu, Taiwan, in 2006. He is currently pursuing the Ph.D. degree in mechanical engineering at Texas A\&M University, College Station, TX, USA.

%     From 2008 to 2009, he was a Research Assistant with the Institute of Physics, Academia Sinica, Tapei, Taiwan. His research interest includes\vadjust{\vfill\pagebreak} the development of surface processing and biological/medical treatment techniques using nonthermal atmospheric pressure plasmas, fundamental study of plasma sources, and fabrication of micro- or nanostructured surfaces. 

%   Mr. Author’s awards and honors include the Frew Fellowship (Australian Academy of Science), the I. I. Rabi Prize (APS), the European Frequency and Time Forum Award, the Carl Zeiss Research Award, the William F. Meggers Award and the Adolph Lomb Medal (OSA).
% \end{IEEEbiography}

\end{document}